%% file: bmvc_camera.tex
\documentclass{bmvc2k}
\usepackage{comment}
\usepackage{amsmath}
\usepackage{amssymb}
\usepackage{booktabs}
\usepackage{multirow}
\usepackage{floatrow}
\newfloatcommand{capbtabbox}{table}[][\FBwidth]
\usepackage{makecell}
\usepackage{wrapfig}
\usepackage{setspace}
\usepackage{xcolor}
\usepackage{color}
\usepackage{caption}

\definecolor{ForestGreen}{RGB}{34,139,34}
\newcolumntype{L}[1]{>{\raggedright\let\newline\\\arraybackslash\hspace{0pt}}m{#1}}
\newcolumntype{C}[1]{>{\centering\let\newline\\\arraybackslash\hspace{0pt}}m{#1}}
\newcolumntype{R}[1]{>{\raggedleft\let\newline\\\arraybackslash\hspace{0pt}}m{#1}}

\title{EBJR: Energy-Based Joint Reasoning for Adaptive Inference}

\addauthor{Mohammad Akbari}{mohammad.akbari@huawei.com}{1}
\addauthor{Amin Banitalebi-Dehkordi}{amin.banitalebi@huawei.com}{1}
\addauthor{Yong Zhang}{yong.zhang3@huawei.com}{2}

\addinstitution{
Huawei Technologies Canada Co., Ltd.
}

\runninghead{Akbari, Banitalebi-Dehkordi, Zhang}{EBJR: Energy-Based Joint Reasoning}

\begin{document}

%\begin{comment}

\maketitle

\begin{spacing}{0.93}

\vspace{-10pt}
\begin{abstract}
State-of-the-art deep learning models have achieved significant performance levels on various benchmarks. However, the excellent performance comes at a cost of inefficient computational cost. Light-weight architectures, on the other hand, achieve moderate accuracies, but at a much more desirable latency. This paper presents a new method of jointly using the large accurate models together with the small fast ones. To this end, we propose an Energy-Based Joint Reasoning (EBJR) framework that adaptively distributes the samples between shallow and deep models to achieve an accuracy close to the deep model, but latency close to the shallow one. Our method is applicable to out-of-the-box pre-trained models as it does not require an architecture change nor re-training. Moreover, it is easy to use and deploy, especially for cloud services. Through a comprehensive set of experiments on different down-stream tasks, we show that our method outperforms strong state-of-the-art approaches with a considerable margin. In addition, we propose specialized EBJR, an extension of our method where we create a smaller specialized side model that performs the target task only partially, but yields an even higher accuracy and faster inference. We verify the strengths of our methods with both theoretical and experimental evaluations. 
% Code and demo are available at \cite{supplementary}.
% the supplementary materials.
\end{abstract}

%-------------------------------------------------------------
\vspace{-13pt}
\section{Introduction}
\label{sec:introduction}
\vspace{-6pt}
Recent years have witnessed exciting achievements in the development of highly capable deep neural networks (DNNs), to the extent that new state-of-the-art (SOTA) results are being published frequently. 
% Recently, tremendous success in the development of highly capable deep neural networks (DNNs) has been witnessed and new state-of-the-art results are being published frequently. 
However, achieving this level of performance requires either using extremely large architectures such as GPT-3 \cite{gpt3} or SEER \cite{seer} with billions of parameters (350GB memory and 175B parameters in GPT-3), or ensembling many models. Consequently, this results in an inefficient inference compared to light-weight models. 

To alleviate the problem of slow inference on large architectures, a natural solution is to apply some form of model compression. Model compression literature is rich and mature, and covers various techniques such as network quantization \cite{hawq, Jacob2018QuantizationAT, zeroq, haq, jin2020neural}, knowledge distillation ~\cite{hinton2015distilling, zhou2000m}, pruning ~\cite{cheng2017survey, he2017channel, gao2020rethinking, le2020paying}, or a combination of multiple techniques ~\cite{polino2018model, cheng2017survey, han2015deep}. After compression, the output DNN may have a reduced number of parameters or may operate in lower bit precision. However, it can be observed that there is a trade-off between the compression ratio and accuracy of a model. Aggressive compression leads to significant performance drop that defeats the purpose. Moreover, compression is one solution for all data and is inference-time deterministic (static) with no flexibility over different data samples. 
% We do not go into further depth on them as they are commonly orthogonal to other approaches in that they can be applied together with the adaptive inference techniques. 
That being said, compression techniques are commonly orthogonal to other approaches in that a degree of compression can be added to other methods.

On the other hand, adaptive inference approaches propose to route to different branches of a DNN either stochastically, or based on some decision criteria on input data \cite{branchynet, blockdrop, msdnet, ranet, yu2018slimmable, yu2019universally, yang2020mutualnet, wang2020resolution}. These methods mostly are based on architecture re-design, i.e., the model needs to be built in a specific way to support dynamic inference. This makes their training more complex and imposes additional non-trivial hyper-parameter tuning. Adaptive inference methods can broadly be categorized as redundancy-based and multi-exit structures. The redundancy-based approaches exploit the parameter redundancy in neural networks. To this end, \cite{lccl} designed a convolution-based controller layer, which reduces the computations in practice, even though increases the overall network size. Or \cite{liu2018dynamic, blockdrop, veit2018convolutional, wang2018skipnet, sact, cnmm} dynamically skip some layers or blocks on the fly via selective layer execution.

%%% multi-exit
Multi-exit or multi-stage approaches, however, are based on architectures in which a network can exit from different paths based on some confidence criteria. Earlier techniques such as BranchyNet \cite{branchynet} incorporated an entropy-based threshold for routing. A similar approach was taken by \cite{panda2016conditional, berestizshevsky2019dynamically} by training side classifiers to navigate to different paths. \cite{msdnet} proposed a multi-scale dense network to reuse feature maps of different scales, which was further improved in \cite{ranet} by designing a resolution adaptive network (RANet) to identify low resolution inputs as easy cases, and process them with cheaper computations. There are also works based on architecture search for dynamic inference models \cite{yuan2020enas4d}. It is also worth noting that the majority of the existing methods focus on the task of image classification and fail to study the other applications. \cite{adaptivefeeding} is an example were adaptive inference was investigated for the task of object detection, by leveraging a Support Vector Machine (SVM) classifier to route the work-load. A down-side for \cite{adaptivefeeding}, however, is that dynamically changing the routing traffic between the fast and slow branches requires retraining.

Although the redundancy-based and multi-exit methods have made a significant progress and work well in practice, we will show that they do not reach the levels of performance provided by our energy-based strategy. In addition, most of these methods require training models in a specific way necessitated by their architecture design. In contrast, our method works with out-of-the-box already trained models without a need for re-training.

\begin{figure*}%[htb!]
    \centering
    \vspace{-4pt}
    \includegraphics[width=1.0\linewidth]{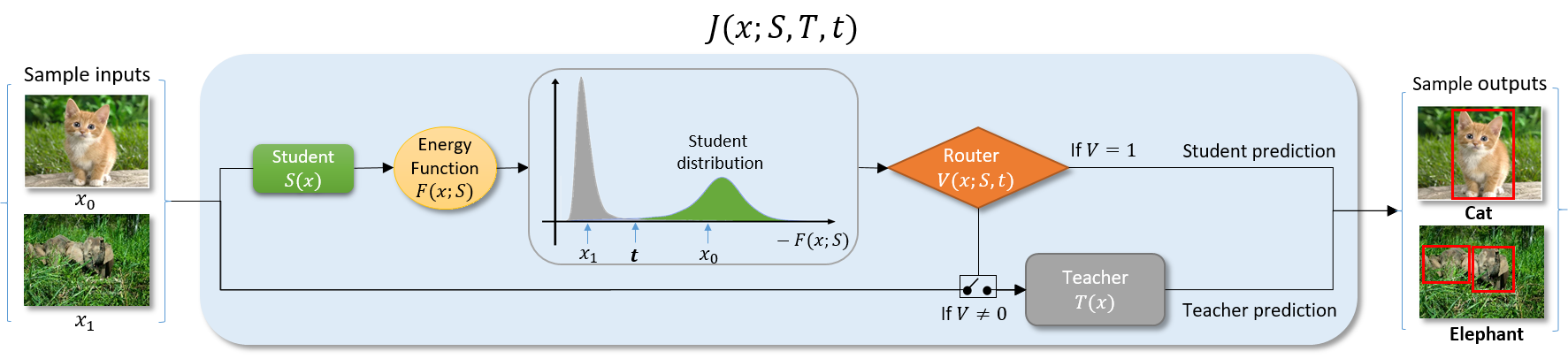}
    \vspace{-15pt}
    \caption{\small The overall flow-diagram of the proposed energy-based joint reasoning (EBJR) method.}
    \label{fig:our-energy-ood}
\end{figure*}

In this paper, we propose an adaptive inference strategy that combines a large, deep accurate model (called Teacher) with a small, shallow fast one (called Student). Our method is based on an effective energy-based decision making module for routing different samples to deep or shallow models. In this way, certain examples will be sent to the Student model that yields high speed inference, and other examples go to the Teacher model, which is slower, but highly accurate. Our method provides an inference-time trade-off between the inference latency and task accuracy. This can be thought of as a knob for the users to play with, and to dynamically choose a desired point in the trade-off based on their required accuracy or latency. Figure \ref{fig:our-energy-ood} shows a high-level schematic of the proposed framework. 

In addition to our main adaptive inference strategy, we provide an extension called specialized EBJR, which provides more accurate and efficient inference by training the Student in a way that it only learns to perform the down-stream task partially (details in Section \ref{ssec:method_jr_specialized}).

The main contributions of this paper are summarized as follows:
\vspace{-1pt}
\begin{itemize}%[leftmargin=*]
    \vspace{-5pt}
    \item Combining small/shallow models (low accuracy and latency) with large/deep models (high accuracy and latency) to achieve \textbf{high accuracy and low latency}. Our method is easy to build and deploy, is architecture agnostic, applicable to different down-stream tasks (e.g., classification and object detection), and can be applied to existing pre-trained models (with no need for re-training).
    \vspace{-8pt}
    \item An \textbf{energy-based routing mechanism} for directing examples to the small (Student) or large (Teacher) models. This allows a dynamic trade-off between accuracy and computational cost that outperforms the previous works in adaptive inference (with zero overhead for real-time adjustment of speed/accuracy). 
    %\vspace{-4pt}
    %\item An \textbf{unsupervised} extension of our method, to facilitate distilling the large Teacher model to small Student model. This is suitable for cases where users provide large models as input, without labeled data, and the goal is to build an efficient inference pipeline.
    \vspace{-8pt}
    \item Creating a small, Student model \textbf{specialized} for a subset of tasks (e.g., top-C classes only) with high accuracy; along with a plus-one (+1) mechanism, to distinguish the top-C-class data from the others.
\end{itemize}

% \begin{itemize}
%     \item LCCL (CVPR2017) \cite{lccl}, SACT (CVPR2017) \cite{sact}, BlockDrop (CVPR2018) \cite{blockdrop}, MSDNet (ICLR2018) \cite{msdnet}, CNMM (ICCV2019) \cite{cnmm}, RANet (CVPR2020) \cite{ranet}.
%     \item Almost all of above works are based on architecture redesign.
%     \item Similar to us, they all propose "Adaptive Inference"-based solutions to deal with the "model efficiency" problem.
% \end{itemize}

\vspace{-15pt}
\section{Energy-Based Joint Reasoning (EBJR)} 
\label{sec:method_jr}
\vspace{-4pt}
We introduce EBJR, a novel energy-based joint reasoning approach for adaptive inference. Our method is inspired by the fact that smaller (shallower/narrower) models typically have lower accuracy, but very fast inference; and larger (deeper/wider) models, on the other hand, are highly accurate, but very slow. We combine the small model (denoted by Student) and the large model (denoted by Teacher) in an efficient and effective way to provide a fast inference, while maintaining the high accuracy. A schematic of our framework is shown in Figure \ref{fig:our-energy-ood}.

The main challenge here is to design an effective routing mechanism (denoted by Router) to decide which model to use for each input. As for the adaptive inference, the Router should also provide the option of dynamic trade-offs between accuracy and latency at the inference time. The Router module essentially operates similar to a binary classifier that directs easy samples to the Student and the hard ones to the Teacher. In some ways, this problem is also similar to the out-of-distribution detection (OOD) problem \cite{liu2020energy} in which in- and out-of-distribution data are differentiated. OOD is generally used when a model sees some input test data that differs from its training data (in-distribution data). Consequently, the predictions of the model on OOD samples would be unreliable. For our case, the Router should be able to identify whether or not the input data fits in the distribution with which the Student has been trained (i.e., there is a high probability that the Student can make accurate predictions for that input data). If not, the data is labelled as hard for the Student and should be forwarded to the Teacher that has higher capability. In our work, we investigate the energy characteristics of data samples to route them effectively.

% Various approaches have been explored to handle OOD problem for both regression and classification tasks. Among them, energy-based methods have recently shown to be very effective, which is the major component for designing our Router in this work. 

\vspace{-10pt}
\paragraph{Energy definitions.}
Given an input data point $\textbf{x}$, the energy function is defined as $E(\textbf{x}): \mathbb{R}-^D \rightarrow \mathbb{R}$ to map the input $\textbf{x}$ to a scalar, non-probabilistic energy value $y$. The probability distribution over a collection of energy values can be defined according to the Gibbs distribution \cite{hinton1994autoencoders, lecun2006tutorial}: $p(y|\textbf{x}) = \frac{1}{Z}\big(e^{-E(\textbf{x},y)}\big),$
% Given an input data point $\textbf{x}$, the energy function $E(x)$ is used to map $x$ to an energy value that is a single, non-probabilistic scalar \cite{hinton1994autoencoders, lecun2006tutorial}: $E(x): \mathbb{R}^D \rightarrow \mathbb{R}$. The probability distribution over a collection of energy values can be defined using the Gibbs distribution: 
where $Z(\textbf{x}) = \int_{y'} e^{-E(\textbf{x},y')}$ is the partition function.
% where $Z$ is the partition function defined as:
% \begin{equation}
% \label{eq:partition_func}
%     Z(\textbf{x}) = \int_{y'} e^{-E(\textbf{x},y')}.
% \end{equation}
The free energy \cite{lecun2006tutorial} of $\textbf{x}$ can then be expressed as the negative log of the partition function:
\vspace{-4pt}
\begin{equation}
\label{eq:free_energy}
    F(\textbf{x}) = -\log \big(Z(\textbf{x})\big).
    \vspace{-4pt}
\end{equation}
In the following subsections, we will describe our energy-based joint reasoning method, and give formulations for classification, regression, and object detection problems.

\vspace{-10pt}
\subsection{Classification}
\label{ssec:method_jr_classification}
\vspace{-3pt}

The Student classifier is defined as a function  $S^c(\cdot)$ for mapping the input $\textbf{x}$ to $C$ real-valued logits (i.e., for $C$ number of class labels): $S^c(\textbf{x}): \mathbb{R}^D \rightarrow \mathbb{R}^C$. In probability theory, we can use the output of the softmax function to represent a categorical distribution that is a probability distribution over $C$ different possible outcomes \cite{liu2020energy}. A categorical distribution using the softmax function is expressed by:
\vspace{-10pt}
\begin{equation}
\label{eq:categorical_dist}
    p(y|\textbf{x}) = \frac{e^{S^c_y(\textbf{x})}}{\sum_i^C e^{S^c_i(\textbf{x})}},
\end{equation}
% \vspace{-8pt}
where $S^c_y (\textbf{x})$ denotes the logit (probability) of the $y$th class label. The energy for a given input $(\textbf{x},y)$ in this case is defined as $E(\textbf{x},y)=-S^c_y(\textbf{x})$ \cite{liu2020energy}. The free energy function $F^c(\textbf{x};S^c)$ is then expressed similar to (\ref{eq:free_energy}) as:
\vspace{-10pt}
\begin{equation}
\label{eq:classifier_free_energy}
    F^c(\textbf{x};S^c) = -\log \sum_i^C e^{S^c_i(\textbf{x})}.
    \vspace{-15pt}
\end{equation}

\vspace{-10pt}
\paragraph{Problem.} We seek to identify samples suitable for the Student and would like to direct the others to the Teacher.
% \paragraph{Formulation.} 
% We propose to use the data density function $p(\textbf{x})$ and mark the data samples with low likelihood as hard (or unfit) for the Student. To this end, the energy-based density function is written as:
A natural solution to this problem is to use the data density function and consider the inputs with low likelihood as hard (or unfit) samples. To this end, an energy-based density function for the Student can be defined as:
\vspace{-4pt}
\begin{equation}
\label{eq:classifier_energy_density}
   p(\textbf{x}) = \frac{1}{Z^c}\big(e^{-F^c(\textbf{x};S^c)}\big),
   \vspace{-4pt}
\end{equation}
where the denominator $Z^c=\int_{\textbf{x}} e^{-F^c(\textbf{x};S^c)}$ is the normalized densities, which can be intractable to compute or estimate. By taking the logarithm of both sides, we obtain:
\vspace{-4pt}
\begin{equation}
\label{eq:classifier_log_density}
    \log \big( p(\textbf{x}) \big) = -F^c(\textbf{x};S^c) - \log(Z^c).
    \vspace{-4pt}
\end{equation}

\vspace{-12pt}
\paragraph{Solution.} The $\log(Z^c)$ term is constant for all $\textbf{x}$, and does not affect the overall energy values distribution. Thus, the negative free energy is linearly aligned with the log likelihood function. This makes it a suitable solution to our problem for detecting easy and hard samples. In this case, higher energy means lower likelihood, which represents harder (or more unfit) samples for the Student's training distribution. 

More precisely, for a threshold $\delta$ on the density function such that $p(\textbf{x}) < \delta$, then a threshold $t$ on the negative free energy can be calculated based on (\ref{eq:classifier_log_density}) as $-F^c(\textbf{x};S^c) < t = \log (\delta Z^c)$.
In practice, for a given input, an energy function is applied to the Student outputs at inference to compute the energy score. Then, if the negative energy value is smaller than a threshold, the input is identified as a bad sample for the Student, and is sent to the Teacher.

% In practice, for a given input, an energy function is applied to the outputs of the Student during inference time to calculate the energy score. Then, if the negative energy value is smaller than a threshold, the input is identified as a bad sample for the Student, and is sent to the Teacher.

%Based on the difference of the energies between in- and out-of-distribution samples.

% add the threshold function (binary classifier -> as our Router) here!
Therefore, given the input data $\textbf{x}$, the Student $S^c(\textbf{x})$, and the threshold $t$, our energy-based Router $V^c(\textbf{x};S^c,t) \in \{0,1\}$ can simply be defined as:
\vspace{-6pt}
\begin{equation}
\label{eq:classifier_Router}
    V^c(\textbf{x};S^c,t) = \begin{cases} \mbox{1} & \mbox{if } - F^c(\textbf{x};S^c) \geq t \\ \mbox{0} & \mbox{if } - F^c(\textbf{x};S^c) < t. \end{cases}
    \vspace{-6pt}
\end{equation}

Let the Teacher classifier be $T^c(\textbf{x}): \mathbb{R}^D \rightarrow \mathbb{R}^{C'}$, with $C = C'$ (the same number of class labels as in the Student). Our joint reasoning classification function $J^c(\textbf{x};S^c,T^c,t) \in [1,C]$ can then be written by: 
\vspace{-10pt}
\begin{equation}
\label{eq:classifier_jr}
    J^c(\textbf{x};S^c,T^c,t) = \begin{cases} \mbox{$S^c(\textbf{x})$} & \mbox{if } V^c(\textbf{x};S^c,t) = 1 \\ \mbox{$T^c(\textbf{x})$} & \mbox{otherwise.} \end{cases}
    \vspace{-6pt}
\end{equation}

\color{black}
\vspace{-11pt}
\subsection{Regression}
\label{ssec:method_jr_regression}
\vspace{-3pt}
A regressor maps an input $\textbf{x}$ to a target scalar $y$ defined as $S^r(\textbf{x}): \mathbb{R}^D \rightarrow \mathbb{R}$. For a given input $(\textbf{x},y)$, the energy function for a regressor is simply defined as $E(\textbf{x},y)=-S^r(\textbf{x},y)$. The regression problem can then be expressed by creating an energy-based model of the conditional density $p(y|\textbf{x})$ as:
\vspace{-14pt}
\begin{equation}
\label{eq:regressor_dist}
    p(y|\textbf{x};S^r) = \frac{e^{S^r(\textbf{x},y)}}{\int_{y'} e^{S^r(\textbf{x},y')}},
\end{equation}
where the denominator is the normalizing partition function that involves a computationally intractable integral. One solution is to obtain its approximations using Monte Carlo importance sampling method as described in \cite{gustafsson2020energy}. The free energy in this case is defined by:
\vspace{-6pt}
\begin{equation}
\label{eq:regressor_free_energy}
    F^r(\textbf{x};S^r)= -\log \left( \int_{y'} e^{S^r(\textbf{x},y')} \right).
    \vspace{-0pt}
\end{equation}

%\vspace{-4pt}
Similar to (\ref{eq:classifier_energy_density}), the density function for a regressor using the energy-based model can be obtained as follows:
\vspace{-4pt}
\begin{equation}
\label{eq:regressor_density}
    p(\textbf{x}) = \frac{1}{Z^r}\big(e^{-F^r(\textbf{x};S^r)}\big),
    \vspace{-0pt}
\end{equation}
where the denominator is the normalized densities defined as $Z^r = \int_{\textbf{x}} e^{-F^r(\textbf{\textbf{x}};S^r)}$. By taking the log of both sides:
\vspace{-4pt}
\begin{equation}
\label{eq:regressor_log_density}
    \log \big( p(\textbf{x}) \big) = -F^r(\textbf{x};S^r) - \log(Z^r),
    %\vspace{-2pt}
\end{equation}
which, as in the classification problem, shows that $-F^r(\textbf{x};S^r)$ has a linear alignment with the log likelihood function by considering the fact that $\log(Z^r)$ is constant for all $\textbf{x}$, which makes it desirable for our problem.

Given the input data $\textbf{x}$, the Student regression model $S^r(\textbf{x})$, and a threshold $t$, our energy-based Router for a regression problem can be simply defined with $V^r(\textbf{x};S^r,t) \in \{0,1\}$ based on (\ref{eq:classifier_Router}). The joint reasoning function $J^r(\textbf{x};S^r,T^r,t) \in \mathbb{R}$ can also be expressed based on (\ref{eq:classifier_jr}), where $T^r(\textbf{x}): \mathbb{R}^D \rightarrow \mathbb{R}$ is the Teacher regression model. 

\vspace{-8pt}
\subsection{Object detection}
\label{ssec:method_jr_od}
\vspace{-3pt}
For the object detection task with a combination of classification and regression, we can define the total free energy score as: $F^o(\textbf{x};S^c,S^r) = F^c(\textbf{x};S^c)+F^r(\textbf{x};S^r)$, where the regressor for predicting 4 points of a bounding box is defined as $S^r(\textbf{x}): \mathbb{R}^D \rightarrow \mathbb{R}^4$. With $B$ number of detected boxes and $C$ labels, the classifier's free energy score $F^c(\textbf{x};S^c)$ is formulated as:
\vspace{-8pt}
\begin{equation}
\label{eq:od_cls_free_energy}
    F^c(\textbf{x};S^c) = \frac{1}{B}\big(-\sum_b^B \log \sum_i^C e^{S^c_{b,i}(\textbf{x})}\big),
    \vspace{-0pt}
\end{equation}
where $S^c_{b,i}$ is the classifier's output for the $i$th class label of the $b$th bounding box with $b \in [1,B]$ and $i \in [1,C]$. The regressor's free energy $F^r(\textbf{x};S^r)$ is also given by:
\vspace{-6pt}
\begin{equation}
\label{eq:od_reg_free_energy}
    F^r(\textbf{x};S^r) = \frac{1}{4B}\big(-\sum_b^B \sum_j^4 \log \int_{y'} e^{S^r_{b,j}(\textbf{x},y')}\big),
    \vspace{-0pt}
\end{equation}
where $S^r_{b,j}$ is the regression output for the $j$th point of the $b$th bounding box with $j \in [1,4]$.

The energy-based joint reasoning function for object detection task is finally defined as:
\vspace{-4pt}
\begin{equation}
\small
    J^o \left(\textbf{x};S^o,T^o,t\right) = \begin{cases} \mbox{$T^o$(\textbf{x})} & \mbox{if } - F^o(\textbf{x};S^o) < t \\ \mbox{$S^o$(\textbf{x})} & \mbox{if } - F^o(\textbf{x};S^o) \geq t, \end{cases}
    \vspace{-4pt}
\end{equation}
where {\small$S^o = \{S^c,S^r\}$ and $T^o = \{T^c,T^r\}$} denote the Student and Teacher object detection models.

%\begin{equation}
%    E(\textbf{x},y) = \frac{1}{2} \lVert g(\textbf{x}) - y \rVert^2,
%\end{equation}

\vspace{-8pt}
\subsection{Specialized EBJR} 
\label{ssec:method_jr_specialized}
\vspace{-2pt}
In Section \ref{ssec:method_jr_classification}, it was assumed that the Student and Teacher models have equal number of classes that is $C = C'$. As proved in \cite{abramovich2019classification}, in order to achieve a good performance for a classifier with large number of classes, significantly large number of features are required. Since the Teacher model is assumed to be a very large model with significant number of features, it is capable of handling more difficult tasks with a large $C'$. On the other hand, the small Student model may lack enough features to be able to effectively deal with a large $C$.

In addition, in inference services such as public clouds, the majority of input data usually belongs to a small, popular subset of classes that are used frequently, for example,``people", ``cat", ``dog", ``car", etc (supplementary materials contain example-per-class histogram plots for public datasets, and confirms this intuition). Considering this fact, the Student can be trained and specialized to be highly accurate on this specific/popular subset (with a small $C$). Consequently, in our joint reasoning scheme, most of the input data can be handled by the Student in a very accurate and computationally efficient way. %In this case, the above-mentioned problem of lacking significant features for large $C$ does not exist anymore. 

Let the specialized Student be $\Bar{S}^c(x): \mathbb{R}^D \rightarrow \mathbb{R}^{\Bar{C}+1}$, where $\Bar{C} \ll C$. To make sure the model can still exploit and learn from all data at the training time, we label the data that do not belong to $\Bar{C}$ as an additional class (i.e., the `other' class). The extra class is also utilized as a supplementary mechanism in our Router to evaluate the performance of $\Bar{S}^c$ on a given input data at the inference time. Similar to a binary classifier, it is used for distinguishing the data with $\Bar{C}$ labels from the others.

The specialized Student $\Bar{S}^c$ has another benefit for our energy-based Router. Since only a subset of class labels is used for training the Student, the energy differences between in- and out-of-distribution data respectively denoted by $(\textbf{x},i)$ and $(\Bar{\textbf{x}},j)$ tends to be larger:
\vspace{-6pt}
\begin{equation}
    | \Bar{F}^c(\textbf{x};\Bar{S}^c_i)-\Bar{F}^c(\Bar{\textbf{x}};\Bar{S}^c_{j}) | > | F^c(\textbf{x};S^c_i)-F^c(\Bar{\textbf{x}};S^c_{j}) |,
    \label{eq:energy_gap}
    \vspace{-0pt}
\end{equation}
%\vspace{-16pt}
where $i \in [1,\Bar{C}]$ and $j \notin [1,\Bar{C}]$. The larger the energy difference, the better the Router can distinguish the fit and unfit data for the Student, which results in more accurate and efficient adaptive inference. Given the input data $\textbf{x}$, the specialized Student $\Bar{S}^c(\textbf{x})$, and a threshold $t$, our specialized energy-based Router $\Bar{V}(\textbf{x};\Bar{S}^c,t) \in \{0,1\}$ is expressed as:
\vspace{-4pt}
\begin{equation}
\small
    \Bar{V}(\textbf{x};\Bar{S}^c,t) = \begin{cases} \mbox{1} & \mbox{if } - \Bar{F}^c(\textbf{x};\Bar{S}^c) \geq t \text{~~ and~~} \Bar{S}^c(\textbf{x}) \in [1,\Bar{C}] \\ \mbox{0} & \mbox{if } - \Bar{F}^c(\textbf{x};\Bar{S}^c) < t \text{~~or~~} \Bar{S}^c(\textbf{x}) \in \{\Bar{C}+1\}, \end{cases}
    \vspace{-4pt}
\end{equation}
where ${\Bar{C}+1}$ denotes the extra class defined in $\Bar{S}^c$. The free energy $\Bar{F}^c(\textbf{x};\Bar{S}^c)$ for the specialized Student is calculated only over the top-$\Bar{C}$ classes, not the extra class, as follows:
\vspace{-10pt}
\begin{equation}
\label{eq:specialized_classifier_free_energy}
    \Bar{F}^c(\textbf{x};\Bar{S}^c) = -\log \sum_i^{\Bar{C}} e^{\Bar{S}^c_i(\textbf{x})} ~~~\text{with}~~~ i \notin \{\Bar{C}+1\}.
    \vspace{0pt}
\end{equation}
%\vspace{-10pt}

Let the Teacher be $T^c(\textbf{x}): \mathbb{R}^D \rightarrow \mathbb{R}^{C'}$ with $\Bar{C} \ll C'$. Then, the specialized joint reasoning function $\Bar{J}(\textbf{x};\Bar{S}^c,T^c,t) \in [1,C']$ for making the predictions related to $\textbf{x}$ can be given by: 
\vspace{-6pt}
\begin{equation}
    \Bar{J}(\textbf{x};\Bar{S}^c,T^c,t) = \begin{cases} \mbox{$\Bar{S}^c(\textbf{x})$} & \mbox{if } \Bar{V}(\textbf{x};\Bar{S},t) = 1 \\ \mbox{$T^c(\textbf{x})$} & \mbox{otherwise}. \end{cases}
\end{equation}
%\vspace{-6pt}

%\begin{equation}
%    F(\textbf{x};\Bar{S}^c) = -log \sum_i^{\Bar{C}} e^{{\Bar{S}}^c_i(\textbf{x})}.
%\end{equation}

% Energy-based Router works better since topN will have different distributions compared to non-topN (the energy differences between topN and non-topN is large).

% +1 helps in two ways: 1) additional help for ood besides the Router, 2) better separation of in- and ood during training (the model is trained with all data and class labels (all features), but also learns both in and out distributions.

%Considering the fact that the majority of input data belongs to a small subset of tasks (classes/categories) (Figure \ref{fig:caltech256-distribution}), our Student can be trained and specialized to be highly accurate on this specific/popular subset. As a result, most of the input data can be handled by the Student and be inferred in a very accurate and computationally efficient way.

%$S^c(\textbf{x}): \mathbb{R}^D \rightarrow \mathbb{R}^{C}$
%$T^c(\textbf{x}): \mathbb{R}^D \rightarrow \mathbb{R}^{C'}$
%where
%$\lvert C' \rvert \ll \lvert C \rvert$

% future work: Use the energy-based OOD (on the Teacher's outputs) for analyzing/choosing the TopN subset of classes for Student. ...

\vspace{-16pt}
\section{Experiments} \label{sec:experiments}
\vspace{-4pt}
%It is important this section follows the methods section in a systematic way. I.e. anything discussed there should be backed up by results here.

%\paragraph{What kind of experiment should we do?} Write the experiments plan/subsections. This way we can understand what's the goal of each experiment (also datasets, methods, etc.)

%\paragraph{For each table/figure, create a pseudo-table or figure, to show the results.}

In this section, we evaluate and discuss the performance of our EBJR approach along with the other related methods on image classification and object detection tasks on different benchmarks. We provide more results and ablation studies in the supplementary materials.

\vspace{-10pt}
\subsection{Adaptive inference results} 
\label{sec:experiments_ai_ic}
\vspace{-4pt}

\begin{figure*}
    \centering
    \begin{subfigure}{0.51\textwidth}
        \centering
        \includegraphics[width=0.99\linewidth]{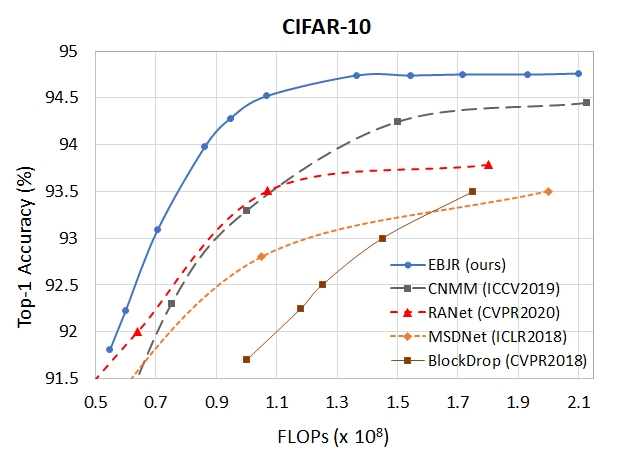}
    \end{subfigure}
    \hspace{-13pt}
    \begin{subfigure}{0.51\textwidth}
        \centering
        \includegraphics[width=0.99\linewidth]{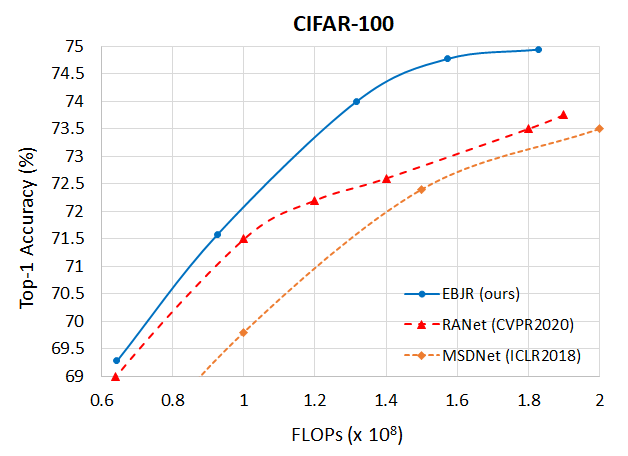}
    \end{subfigure}
    \vspace{-10pt}
    \caption{\label{fig:ic_comparison_results} \small Evaluation of EBJR and the SOTA in adaptive inference on CIFAR datasets.}
\end{figure*}

\begin{figure*}
    \centering
    \vspace{-4pt}
    \begin{subfigure}{0.50\textwidth}
        \centering
        \includegraphics[width=0.99\linewidth]{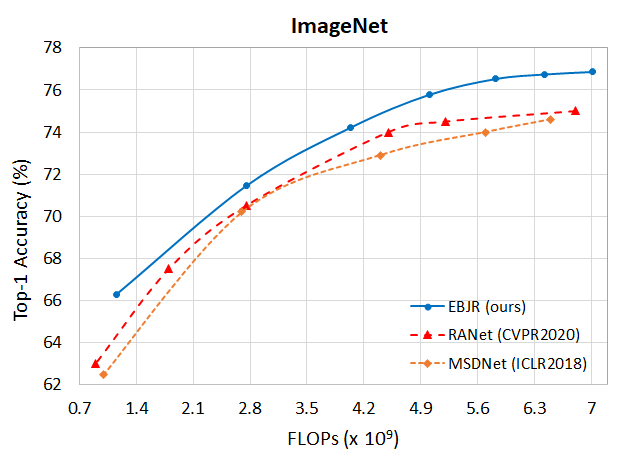}
    \end{subfigure}
    \hspace{-10pt}
    \begin{subfigure}{0.50\textwidth}
        \centering
        \includegraphics[width=0.99\linewidth]{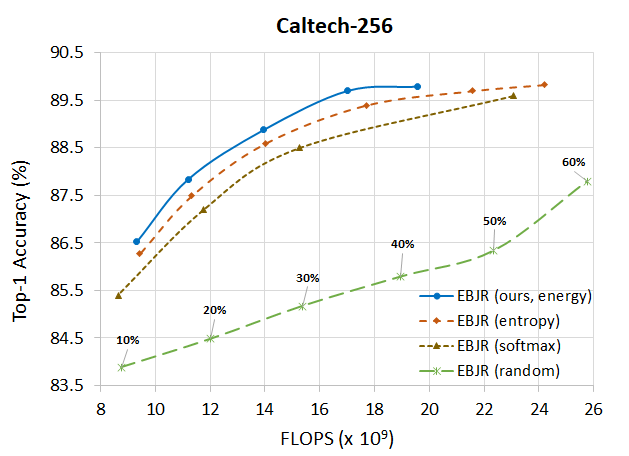}
    \end{subfigure}
    \vspace{-5pt}
    \caption{\small Evaluation of EBJR on the ImageNet (left) and Caltech-256 (right) datasets. {The numbers on the EBJR (random) curve show the percentage of samples processed by the Teacher.}}
    \label{fig:ic_comparison_results_imagenet}
\end{figure*}

Figures \ref{fig:ic_comparison_results} and \ref{fig:ic_comparison_results_imagenet} show the classification results for EBJR and the SOTA in adaptive inference on CIFAR-10, CIFAR-100, ImageNet, and Caltech-256 \cite{caltech256} datasets. We use multiple datasets not only to evaluate the generality of our method, but also because not all other methods published results on a single standard dataset. For all the datasets, we use DenseNet models \cite{densenet} for our Student and Teacher, except Caltech-256 for which ResNet models are used. Table \ref{tbl:ic_models} shows and compares the details about the Student, Teacher, and EBJR models and their accuracy, floating point operations (FLOPs), {and average inference time (latency).} 

Note that many previous approaches are based on the DenseNet architecture, and then adaptively dropping connections for inference speed-up. Thus, we also choose DenseNet as the main architecture to establish a fair comparison although our method does not rely on any specific network design and can work with any black-box architectures. Moreover, 
%since different methods deviate slightly from the DenseNet-like architectures according to their adaptive nature, 
we follow the standard practice in the previous works and analyze the results with FLOPs \cite{ranet,msdnet,cnmm,blockdrop}. For our method, the total FLOPs is measured as a weighted average of the Teacher and Student FLOPs based on their usage frequency as:
$FLOPs=\frac{1}{N_S+N_T}\big(N_{S}.F_S+N_{T}.(F_S+F_T)\big)$, where $N_S$ and $N_T$ are respectively the number of samples processed by Student (with $F_S$ FLOPs) and Teacher (with $F_T$ FLOPs). Note that the metric used in \cite{ranet,msdnet,cnmm} is multiply-accumulates (MACs), i.e., half the FLOPs used in this work.

In Figures \ref{fig:ic_comparison_results} and \ref{fig:ic_comparison_results_imagenet}, the trade-off between accuracy and computational cost is adaptively achieved in our method by choosing different values for the threshold parameter $t$ as defined in (\ref{eq:classifier_Router}) and (\ref{eq:classifier_jr}). {The larger the threshold, the more input data are routed to the Teacher model, which results in more accurate, but slower inference. As the Student is able to make accurate predictions for the majority of input data, the adaptive inference with an appropriate small enough $t$ can almost reach the Teacher's accuracy but with a much lower computational cost.
For CIFAR-10, this strategy achieves the Teacher's accuracy with $\approx$2.2$\times$ less FLOPs. It can also lead to approximately 3$\times$ less FLOPs with an accuracy of $\approx$94.5\% (i.e., only $\approx$0.2\% lower than the Teacher)}. The amount of speed-up for Caltech-256 is about 2$\times$, while maintaining the Teacher's accuracy of 89.87\%. For CIFAR-100 and ImageNet, which are more complicated benchmarks, Teacher's top-1 accuracy is almost achieved with approximately 1.5$\times$ savings on the computations.
Moreover, as illustrated in Figures \ref{fig:ic_comparison_results} and \ref{fig:ic_comparison_results_imagenet}, our method outperforms the previous works such as RANet \cite{ranet} and MSDNet \cite{msdnet} on all the three benchmarks across a variety of accuracy and cost combinations.

\begin{table*}
\vspace{-2pt}
\fontsize{6.5}{9}\selectfont
\begin{center}
\begin{tabular}[t]{p{1.12cm}p{0.4cm}p{0.4cm}cp{0.4cm}p{0.4cm}cp{0.4cm}p{0.4cm}cp{0.4cm}p{0.4cm}c}
\toprule
\multicolumn{1}{c}{ } & \multicolumn{3}{c}{\textbf{CIFAR-10}} & \multicolumn{3}{c}{\textbf{CIFAR-100}} & \multicolumn{3}{c}{\textbf{ImageNet}} &
\multicolumn{3}{c}{\textbf{Caltech-256}}\\
\\[-0.30cm]
\cmidrule(l{3pt}r{3pt}){2-4} \cmidrule(l{3pt}r{3pt}){5-7} \cmidrule(l{3pt}r{3pt}){8-10} \cmidrule(l{3pt}r{3pt}){11-13} 
\\[-0.30cm]
& \textbf{S} & \textbf{T} & \textbf{EBJR} & \textbf{S} & \textbf{T} & \textbf{EBJR} & \textbf{S} & \textbf{T} & \textbf{EBJR} & \textbf{S} & \textbf{T} & \textbf{EBJR} \\
\\[-0.30cm]
\midrule
\hspace{-5pt}\textbf{Depth} & 52 & 64 & - & 58 & 88 & - & 121 & 201 & - & 18 & 152 & -\\
\\[-0.30cm]
\hspace{-5pt}\textbf{Growth Rate} & 6 & 12 & - & 6 & 8 & - & 12 & 32 & - & - & - & - \\
\\[-0.30cm]
\hspace{-5pt}\textbf{Accuracy} {\tiny{($\%$)}} & 91.81 & 94.76 & \textbf{94.74} & 69.28 & 74.94 & \textbf{74.87} & 66.28 & 76.92 & \textbf{76.62} & 83.16 & 89.87 & \textbf{89.87} \\
\\[-0.30cm]
\hspace{-5pt}\textbf{FLOPs} {\tiny{($\times10^8$)}} & 0.54 & 2.92 & \textbf{1.36} & 0.64 & 2.14 & \textbf{1.57} & 11.51 & 86.37 & \textbf{58.1} & 54.0 & 340.0 & \textbf{170.1} \\
\\[-0.30cm]
{
\hspace{-5pt}\textbf{Latency} {\tiny{(ms)}}
}
& 14.0 & 35.0 & \textbf{23.78} & 26.0 & 51.0 & \textbf{42.1} & 84.0 & 225.0 & \textbf{196.8} & 25.0 & 200.0 & \textbf{113.6} 
\\
\\[-0.30cm]
\bottomrule
\end{tabular}
\end{center}
\vspace{-16pt}
\caption{\label{tbl:ic_models}\small {Comparison of EBJR} with the Student (\textbf{S}) and Teacher (\textbf{T}) DensetNet models for CIFAR-10, CIFAR-100, and ImageNet; and ResNet models for Caltech-256 experiments.}
\end{table*}

\begin{figure*}
    \centering
    \begin{subfigure}{0.32\textwidth}
        \centering
        \includegraphics[width=0.99\linewidth]{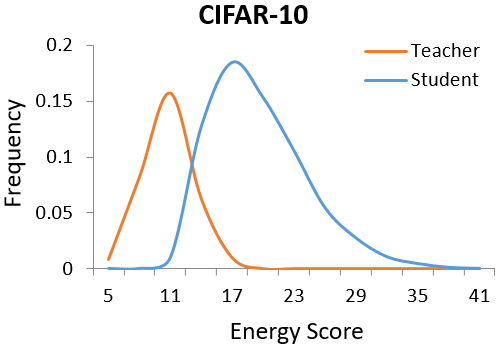}
    \end{subfigure}
    \begin{subfigure}{0.32\textwidth}
        \centering
        \includegraphics[width=0.99\linewidth]{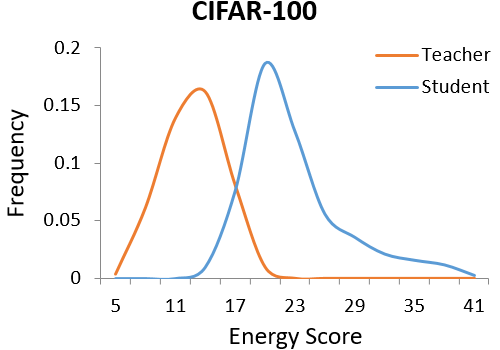}
    \end{subfigure}
    \begin{subfigure}{0.32\textwidth}
        \centering
        \includegraphics[width=0.99\linewidth]{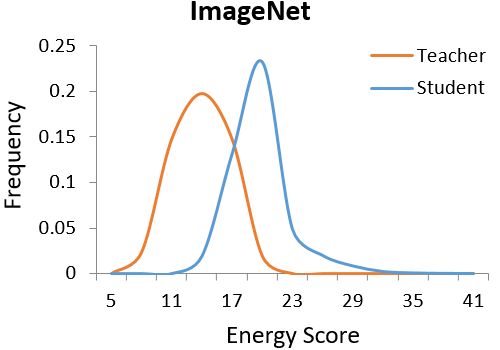}
    \end{subfigure}    
    \vspace{-5pt}
    \caption{\small Energy score distribution for CIFAR-10, CIFAR-100, and ImageNet.}
    \label{fig:energy_distribution}
\end{figure*}

%In addition to energy, softmax and entropy \cite{branchynet} scores can also be used for analyzing the Student's performance in the routing mechanism described in Section \ref{sec:method_jr}. 
To investigate the performance of energy-based routing mechanism compared to other alternatives, we perform an ablation study on Caltech-256, where the energy score is replaced by the softmax confidence or entropy \cite{branchynet} scores. {We also include the random baseline in this experiment, where the input samples are randomly distributed between the Student and Teacher models (the experiment was run multiple times and the best of them was reported).}
The corresponding adaptive inference results are presented in Figure \ref{fig:ic_comparison_results_imagenet}-right. It is observed that softmax- and entropy-based mechanisms can reach the Teacher's accuracy with $\approx$1.4$\times$ and $\approx$1.7$\times$ less FLOPs, which is lower than the energy-based strategy with 2$\times$ speed-up. The theoretical analysis for the entropy score will be given in the supplementary materials.

%that respectively is 12.0, 15.0, and 15.5 for CIFAR-10, CIFAR-100, ImageNet. For CIFAR-10, 

%the experiments given in Figures 2 and 3. For CIFAR-10, 
%Student Accuracy: 0.99, t=12.0 (~70%), CIFAR-10
%Student Accuracy: 0.92, t=15.0 (~55%), CIFAR-100
%Student Accuracy: 0.90, t=15.5 (~45%), ImageNet

%The optimal thresholds are as follows (which can also be observed from the figure):
%* CIFAR-10:  12.0 (~70% exit rate, and Student's accuracy 99%)
%* CIFAR-100: 15.0 (~55% exit rate, and Student's accuracy 92%)
%* ImageNet:  15.5 (~45% exit rate, and Student's accuracy 90%)

Figure \ref{fig:energy_distribution} illustrates the energy score distribution for the samples processed by the Student (i.e., in-distribution data) and Teacher (i.e., out-of-distribution data). As observed, the in-distribution samples (suitable for the Student) tend to have higher energy scores. Based on our experiments, the optimal setup for EBJR is achieved by choosing the threshold $t$ at the crossing point of the two distributions. As a consequence, by choosing $t$=12.0 for CIFAR-10, $\approx$70\% of the samples are handled by the Student with an accuracy of 99.0\%, and only $\approx$30\% are routed to the Teacher, which results in $\approx$3X less total FLOPs. For CIFAR-100 (with $t$=15.0) and ImageNet (with $t$=15.5), $\approx$50\% are processed by the Student (with an accuracy of $\approx$91.0\%), which achieve about 1.5X less FLOPs.

\begin{wraptable}[5]{r}{0.55\textwidth}
\centering
\setlength{\tabcolsep}{5pt}
\fontsize{6.5}{8}\selectfont
\begin{tabular}[t]{p{1.55cm}cccc}
\toprule
\multicolumn{1}{c}{ } & 
\multicolumn{2}{c}{\textbf{CIFAR-10}} & 
\multicolumn{2}{c}{\textbf{ImageNet}}\\ 
\cmidrule(l{3pt}r{3pt}){2-3} \cmidrule(l{3pt}r{3pt}){4-5}
& \textbf{EBJR} & \cite{tann2016runtime} & \textbf{EBJR} & \textbf{BL-Net}~\cite{park2015big} \\
\midrule
\hspace{-4pt}\textbf{Accuracy loss} {\tiny{($\%$)}} & \textbf{0.0} & 0.96 & \textbf{0.9} (0.0) & 0.9 \\
\hspace{-4pt}\textbf{Power savings} {\tiny{($\%$)}} & \textbf{64.03} & 58.74 & \textbf{56.63} (32.93) & 53.7 \\
\bottomrule
\end{tabular}
\vspace{-6pt}
\caption{\label{tbl:comparison_results_power}\small Power consumption vs. accuracy comparison.}
\end{wraptable}

\begin{figure*}
    \centering
    \begin{subfigure}{0.50\textwidth}
        \centering
        \includegraphics[width=0.99\linewidth]{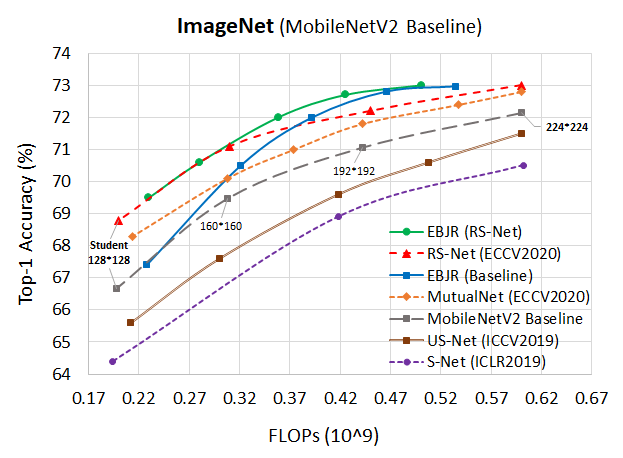}
    \end{subfigure}
    \hspace{-10pt}
    \begin{subfigure}{0.50\textwidth}
        \centering
    \includegraphics[width=0.99\textwidth]{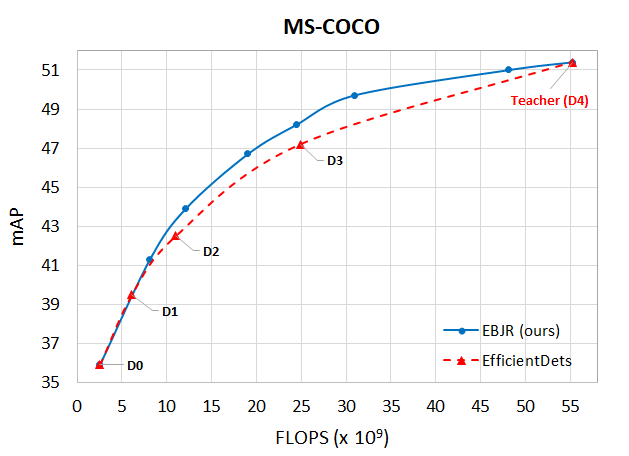}
    \end{subfigure}
    \vspace{-5pt}
    \caption{\small {\textbf{(Left)} Evaluation of MobileNetV2-based EBJR with previous works on ImageNet.} \small \textbf{(Right)} The performance of EBJR for object detection on MS-COCO (compared with EfficientDet \cite{efficientdet}).}
    \label{fig:ic_comparison_results_imagenet_mobilenet_coco}
\end{figure*}

\vspace{-13pt}
\paragraph{Power-Accuracy Tradeoff.}
In the literature, there are also some adaptive inference methods that are proposed for efficient power-accuracy trade-off, for example, \cite{tann2016runtime} and BL-Net \cite{park2015big}. In order to compare EBJR with these approaches, we use the strategy in \cite{lee2019energy} to calculate the power (or energy) consumption per image. As summarized in Table \ref{tbl:comparison_results_power}, the method in \cite{tann2016runtime} reduces the power consumption by 58.74\% with 0.96\% accuracy loss on CIFAR-10, while EBJR (Figure \ref{fig:ic_comparison_results}) achieves 64.03\% power savings without any accuracy loss. Moreover, BL-Net \cite{park2015big} achieves 53.7\% reduction in power consumption with an accuracy loss of 0.9\% on ImageNet. EBJR (Figure \ref{fig:ic_comparison_results_imagenet_mobilenet_coco}), on the other hand, provides a reduction of 56.63\% in power consumption with the same accuracy drop. Unlike BL-Net that does not reach the big model's accuracy, our method achieves the Teacher's accuracy with 32.93\% less power consumption.

\vspace{-12pt}
\paragraph{MobileNetV2-Based EBJR.}
In addition to DenseNet, there exist some SOTA that are based on other architectures such as MobileNetV2 \cite{sandler2018mobilenetv2}, for example, S-Net \cite{yu2018slimmable}, US-Net \cite{yu2019universally}, Mutual-Net \cite{yang2020mutualnet}, and RS-Net \cite{wang2020resolution}. In order to compare EBJR with these approaches, we run another set of experiments on ImageNet, where MobileNetV2 models with 128$\times$128 and 224$\times$224 input resolutions are respectively used as our Student and Teacher.
As shown in Figure \ref{fig:ic_comparison_results_imagenet_mobilenet_coco}-Left, EBJR achieves better performance than S-Net, US-Net, and MutualNet across all FLOPs, and also better than RS-Net in high FLOPs. RS-Net provides better results than EBJR in low FLOPs, which is due to the less accurate Student used in EBJR. However, when EBJR and RS-Net are integrated and the RS-Net's 128$\times$128 path is employed as the Student, the results are improved and EBJR outperforms RS-Net at all trade-off points.

\begin{comment}
\begin{figure*}
    \centering
    \begin{subfigure}{0.50\textwidth}
        \centering
        \includegraphics[width=0.99\linewidth]{figures/imagenet-comparison-mobilenet.png}
    \end{subfigure}
    \vspace{-18pt}
    \caption{\small {Evaluation of MobileNetV2-based EBJR with previous works on ImageNet}}
    \label{fig:ic_comparison_results_imagenet_mobilenet}
\end{figure*}
\end{comment}

%EBJR(densenet) (cifar10):
%64.03\% reductions in energy consumption
%0.0\% accuracy loss
%---
%81.23\% reductions in energy consumption
%0.8\% accuracy loss

%EBJR(mobilenetv2) (imagenet):
%32.93\% reductions in energy consumption
%0.0\% accuracy loss
%---
%56.63\% reductions in energy consumption
%0.9\% accuracy loss

%runtime-Net (cifar10):
%58.74\% reductions in energy consumption
%0.96\% accuracy loss

%BL-Net (imagenet):
%53.7\% reductions in energy consumption
%0.9\% accuracy loss

% \cite{lee2019energy}: reference for energy consumption calculation

\color{black}

\begin{comment}
\begin{wrapfigure}{r}{0.48\textwidth}
  \vspace{-16pt}
  \begin{center}
    \hspace{-10pt}
    \includegraphics[width=0.99\textwidth]{figures/coco-comparison-v2.png}
  \end{center}
  \vspace{-20pt}
  \caption{\small The adaptive inference performance of EBJR for object detection on MS-COCO (compared with the EfficientDet models \cite{efficientdet}).}
  \label{fig:od_comparison}
\end{wrapfigure}
\end{comment}

{

\vspace{-15pt}
\paragraph{Significance Test.}
In order to evaluate the statistical significance of the results, we perform the McNemar's test \cite{dietterich1998approximate} over EBJR and SOTA including RANet and RS-Net. The McNemar's test is interpreted based on a given significance level $\alpha$ (commonly set to 0.05 showing 95\% confidence) as well as the $p$-value and odds ratio calculated by the test.
The default assumption (null hypothesis), i.e., if $p > \alpha$, 
states that the two classifiers should have the same error rate or there should be no difference in the disagreements between them. However, if null hypothesis is rejected, i.e., if $p \leq \alpha$, it suggests that the two classifiers disagree in different ways. After running the test over the EBJR vs. RANet predictions on CIFAR-10 and the EBJR vs. RS-Net predictions on ImageNet
%at $1.1\times10^8$ FLOPs
, $p$-values of $3.2\times10^{-5}$ and $2.4\times10^{-4}$ are respectively obtained. The very low p-values ($\ll 0.05$), which reject the null hypothesis, strongly confirms that there is a significant difference in the disagreements between EBJR and other two models. Also, an odds ratio of 1.42 and 1.14 is respectively obtained, which gives an estimation of how much better EBJR is compared to RANet and RS-Net.
}

Unlike image classification, the adaptive inference for the object detection task has rarely been explored. We analyze the performance of EBJR on the task of object detection (formulated and described in Section \ref{ssec:method_jr_od}) on the MS-COCO dataset \cite{coco}. We employ the EfficientDet-D0 and EfficientDet-D4 \cite{efficientdet} as the Student and Teacher, respectively. %For the unsupervised EBJR, the OID training set is used as the unlabeled set. % we can explained how it has been prepared for our experiment.
%Table \ref{tbl:od_models} reports the performance of Student trained in the supervised setting, compared to Teacher. %For the unsupervised case, we tested different amounts of unlabeled data from OID. We observe that when sufficient unlabeled data (e.g., 1.7M) are provided, the unsupervised Student can perform even better than the supervised one.
Figure \ref{fig:ic_comparison_results_imagenet_mobilenet_coco}-Right shows the adaptive inference results compared to the EfficientDet models (D0, D1, D2, D3, and D4). As shown in the figure, EBJR outperforms the standard EfficientDet models, where it reaches 97\% of the Teacher's mAP on MS-COCO with 1.8$\times$ speed-up. For the same mAP level, the adaptive feeding method of \cite{adaptivefeeding} reports only 1.3$\times$ speed-up.

% Compared to the adaptive feeding method of \cite{adaptivefeeding}, EBJR reaches 97\% of the Teacher's mAP on COCO at 1.8$\times$ speed-up, while \cite{adaptivefeeding} reports that mAP at only 1.3$\times$ speed-up.

% The adaptive feeding method in \cite{adaptivefeeding} is reported to achieve 97\% of their large model's accuracy on COCO, but 1.31$\times$ faster. Our method, however, achieves the same performance with 1.78$\times$ faster inference compared to our large model.

%theirs (on coco test-dev)
% 97\% accuracy of teacher (25/19) - 1.31 times faster than teacher
%----
%ours (coco)
% 97\% accuracy of teacher - with 1.78 times faster than teacher

\vspace{-8pt}
\subsection{Specialized EBJR} 
\label{ssec:experiments_sai_ic}
\vspace{-2pt}
In Section \ref{ssec:method_jr_specialized}, we argued that creating a specialized Student targeted to handle only the popular categories can make the joint inference more efficient. To study this case we run a set of experiments on a subset of the Open Images dataset (OID)~\cite{oid} that has been labeled using the 256 class labels of Caltech-256 dataset. We train the Student with 20\% of the class labels (i.e. $\Bar{C}$=50 out of 256 labels) along with an extra one reserved for the other classes. In this setup, we choose the top-50 class labels with the most number of samples in OID training set. For testing, we randomly select a new set of size 3K from the OID validation set, where 75\% of the data have the top-50 of the labels. This is done to ensure the initial assumption of `having the majority of samples from the popular classes' remains valid.

\begin{figure*}
    \centering
    \begin{subfigure}{0.50\textwidth}
    \centering
        \includegraphics[width=0.99\linewidth]{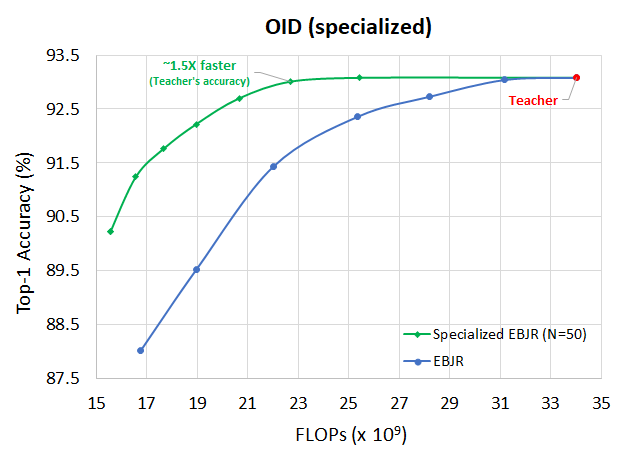}
        \label{fig:caltech256-oid-specialized-specialized}
        % \caption{}
    \end{subfigure}
    \hspace{-10pt}
    \begin{subfigure}{0.50\textwidth}
    \centering
        \includegraphics[width=0.99\linewidth]{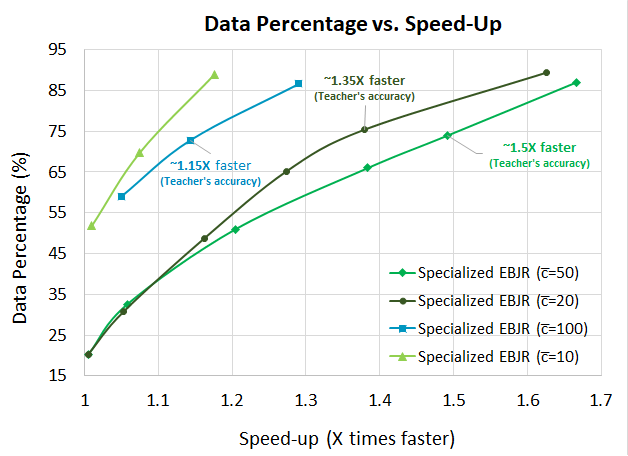}
        \label{fig:caltech256-oid-specialized-data}
        % \caption{}
    \end{subfigure}
    \vspace{-15pt}
    \caption{\small The performance of the specialized EBJR on OID validation set. (left) The specialized EBJR with $\Bar{C}=50$ compared with the general case. (right) The impact of different input data percentages with different chosen subsets of classes for specialized EBJR.}
    \label{fig:caltech256-oid-specialized}
\end{figure*}

Figure \ref{fig:caltech256-oid-specialized}-left shows the results of this experiment. We see that compared to the general cases of EBJR, the specialized EBJR provides the best performance under the assumption that the majority of input data belong to a small subset of classes. For example, compared to the Teacher, the specialized EBJR achieves $\approx$1.5$\times$ less FLOPs with the same accuracy.

Figure \ref{fig:caltech256-oid-specialized}-right shows the effect of the percentage of data that belong to the top-$\Bar{C}$ classes. As expected, the more data in the top-$\Bar{C}$ classes, the faster the joint model, since more load will be directed to the Student which is faster than the Teacher. We observe that when $\Bar{C}$ is too low or too high, e.g., $\Bar{C}$=10 or 100, the adaptive inference with the specialized Student becomes less efficient even with large percentages of data in the top-$\Bar{C}$ classes. For $\Bar{C}=20$ or 50, the specialized EBJR becomes more efficient, especially when $50\%$ or more of data belong to top-$\Bar{C}$ classes. More analysis will be given in the supplementary materials.
% $\Bar{C}=50$ is shown to be more efficient than the other cases across a variety of data percentages.

% \mo{it is not clear here what configuration and models we have used for the Student and Teacher. I think we need to make it clear. In addition, This is the first time we use "gating" term. I think it would be better to follow the previous terms used such as "routing" or "Router"}

% In Table \ref{tbl:caltech_different_rounters}, the performance of our energy-based Router compared with other mechanisms such as entropy, softmax (see Section \ref{ssec:method_jr_confidence}), and random scores is given, which shows how fast (or efficient) the Teacher's accuracy can be reached with these mechanisms. This ablation study is performed on supervised EBJR and Caltech-256 (see Section \ref{ssec:experiments_sai_ic} for more details). %For the random score, the experiment was run multiple times and the best of them was reported. 
% As observed, the energy score provides the most effective and efficient routing mechanism compared to the others. 
% %\mo{we can provide more details in the supplementary: for example, how the entropy and random score is obtained. we can also add the figure we discussed last night.}

{
Note that EBJR is orthogonal to SOTA dynamic inference approaches, including the weight-sharing ones. In Figure 5-left, we applied EBJR on RS-Net, and showed an improved performance on top of it. More results are given in the supplementary materials.}

{
One limitation of EBJR is memory overheard due to the need of both Student and Teacher at inference time. One solution to deal with this problem is to perform the largest possible Student on the edge, but the Teacher on the cloud. If a desired accuracy on the edge is not met, the Router sends certain samples to the cloud for higher accuracy. Since Student is the largest size that can fit to the device, it is expected to handle most cases, while cloud will be used only sparingly in accuracy-sensitive applications. In this setup, the overall accuracy is not bounded by what can run on edge, but the upper-bound is what can run on cloud.
}

\vspace{-10pt}
\section{Conclusion} \label{sec:conclusion}
\vspace{-4pt}
In this paper, we presented an adaptive inference method that combines large, but accurate models with small, but fast models. We proposed an effective energy-based routing module for directing different samples to deep or shallow models. Our method provided a trade-off between the inference latency and accuracy, which in practice is a useful knob for the users to adjust based on their required accuracy or latency, without a need for re-training. In addition, we provided an extension to our method for increasing the inference efficiency by training the shallow models in a way that they only learn to perform the down-stream tasks partially. We presented theoretical and experimental evaluations in support of our method. We hope our work can help facilitate building efficient multi-model inference systems.
%\balance

\end{spacing}

\clearpage
{\small
\bibliography{egbib}
}

%\end{comment}

%\appendix
\clearpage
\input{supplementary.tex}

\begin{comment}
\section{To-Do Items:}
\begin{itemize}
    \item \textbf{1st draft done.} Writing the rebuttal letter.
    \item \textbf{Done.} Add running-time info to the results.
    \item \textbf{Done.} Add Random routing results.
    \item \textbf{Done.} Move energy vs. softmax analysis from sup. to the main body.
    \item \textbf{Done.} EBJR with MobileNet-V2 on ImageNet to be compared with S-Net \cite{yu2018slimmable}, US-Net \cite{yu2019universally}, and Mutual-Net \cite{yang2020mutualnet} (also RS-NET \cite{wang2020resolution}).
    \item \textbf{Done.} Reviewers have also requested us to compare with \cite{park2015big} and \cite{tann2016runtime}, but these are pretty old papers with old architectures such as VGG and Alex.
    \item \textbf{Done.} Statistical significance test.
    \begin{itemize}
        \item Reviewer 3: Providing a statistical significance test (or even reporting variance for a few experiments set up like CIFAR-10 image classification) would be useful to ensure that the results are statistically significant.
    \end{itemize}
    \item Add a small table to include FlOP/time of EBJR when acc is intentionally allowed to drop a bit (0-1-2-5-..)
\end{itemize}
\end{comment}

\end{document}

%% file: supplementary.tex
\section{Supplementary materials} \label{sec:supplementary}

This section contains the supplementary materials.

\subsection{Demo}

% \subsection{Code and demo}
% \subsubsection*{Source code}
% \vspace{-6pt}
% We share our implementation code to make it easy to reproduce our results. The source-code is attached to the supplementary materials in the `code' directory. We also provide detailed instructions for training and evaluating our models in the `README.md' files.

% \vspace{40pt}

% \begin{minipage}{\linewidth}
\begin{figure*}[!b]
\centering\includegraphics[width=1.0\columnwidth]{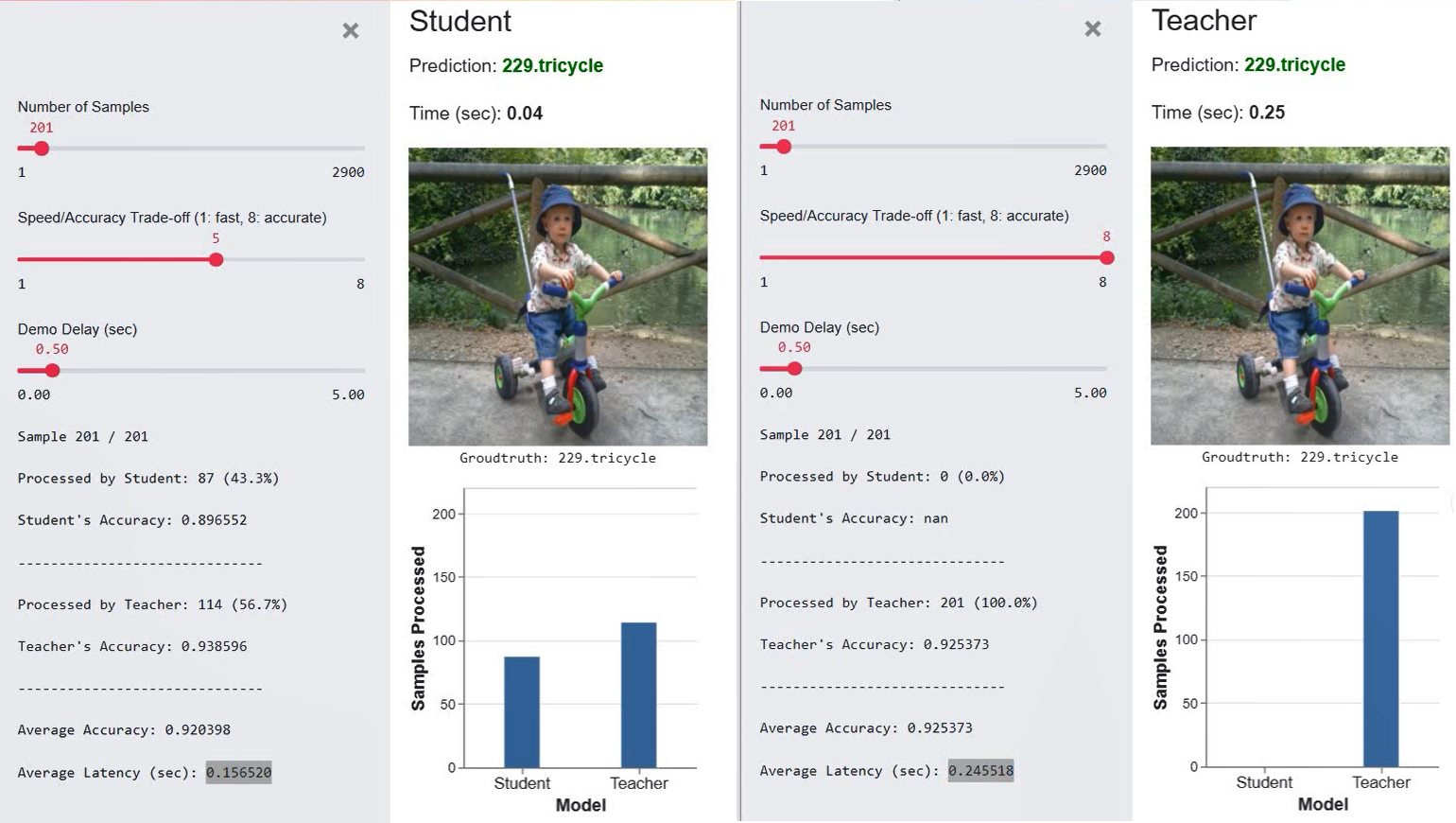}
% \vspace{-12pt}
\caption{\small A screenshot of the provided demo application.}
\label{supp:demo_screenshot}
\end{figure*}
% \end{minipage}

% \subsubsection*{Demo}
\vspace{-6pt}
In addition to the code, we also include a `Demo.mp4' video file that contains a demonstration of our framework. This is based on screen recording of a web application we built to showcase the use-cases of our method in real-world scenarios. Figure \ref{supp:demo_screenshot} shows a screenshot of the demo application.

\subsection{Ablation studies on CIFAR-10, CIFAR-100, and ImageNet}

\begin{figure*}[!htb]
    \centering
    \includegraphics[width=0.49\linewidth]{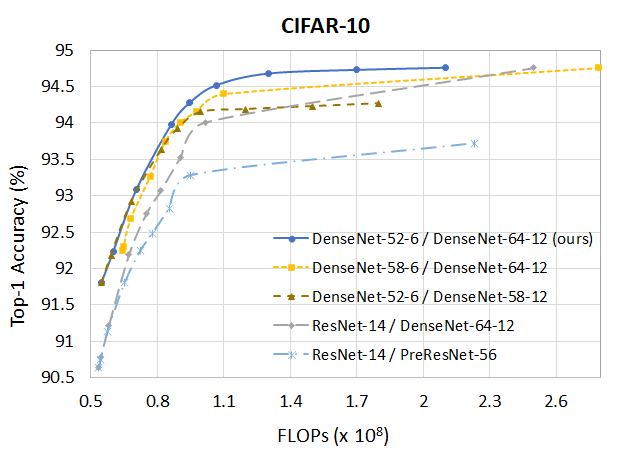}
    \includegraphics[width=0.49\linewidth]{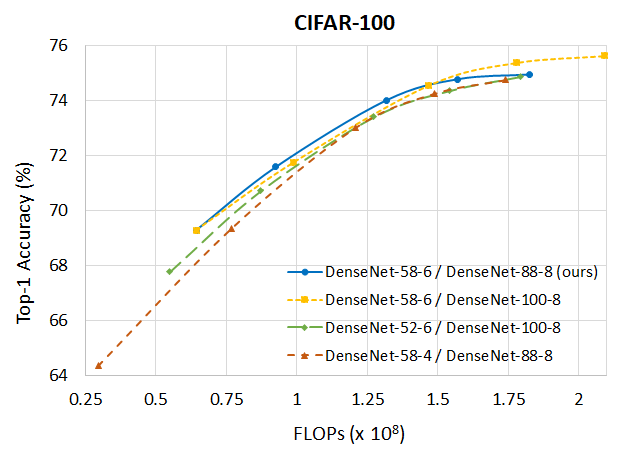}
    \includegraphics[width=0.49\linewidth]{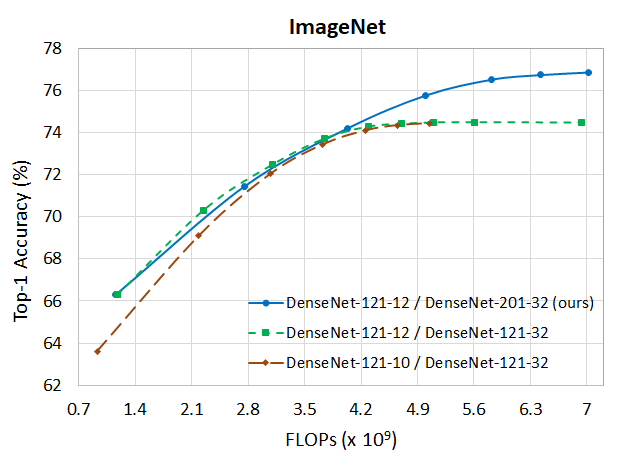}
    \vspace{-4pt}
    \caption{Ablation studies of EBJR on CIFAR-10, CIFAR-100, and ImageNet with different architectures for the Student and Teacher models. DenseNet-$d$-$g$ denotes a densenet model with depth of $d$ and growth rate of $g$. ResNet-$d$ denotes a resnet model with depth of $d$.}
    \label{fig:ablation_studies_cifar}
\end{figure*}

Figure \ref{fig:ablation_studies_cifar} shows the results of ablation studies of our EBJR method with different architectures for Student and Teacher models on CIFAR (10 and 100) and ImageNet. We observe that the results do not vary excessively, which shows the robustness of the proposed method.

\subsection{More experiments with RANet}

\begin{figure*}[!htb]
    \centering
    \includegraphics[width=0.49\linewidth]{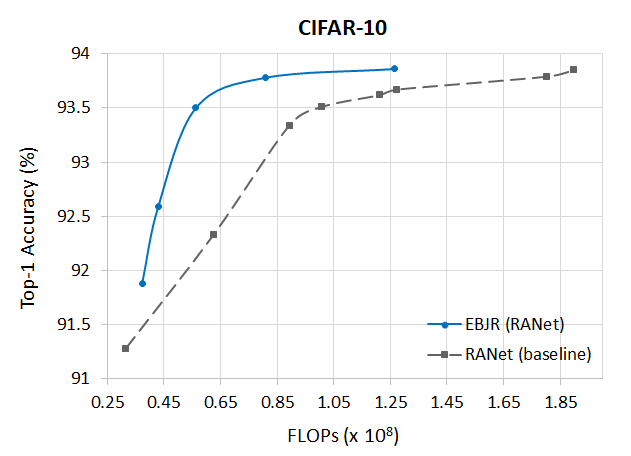}
    \includegraphics[width=0.49\linewidth]{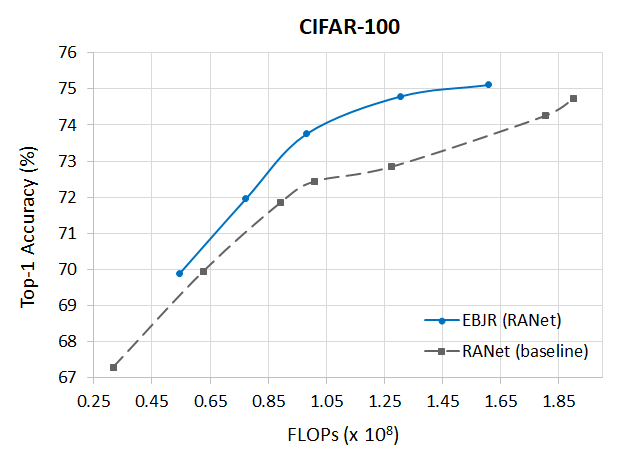}
    \includegraphics[width=0.49\linewidth]{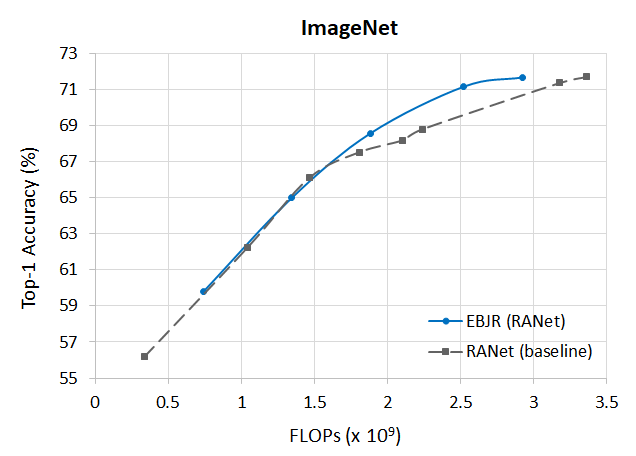}    
    \vspace{-8pt}
    \caption{The performance of EBJR with RANet architecture, compared to the baseline RANet on CIFAR-10, CIFAR-100, and ImageNet.}
    \label{fig:ebjr_ranet}
\end{figure*}

In this experiment, we evaluate the performance of EBJR when the SOTA architectures are used as our Student and Teacher models. In other words, we investigate whether our method can be added on top of other efficient methods such as RANet to benefit both from their designs and our joint inference. To this end, we trained the RANet architecture with three scales (as suggested in RANet work) on CIFAR-10, CIFAR-100, ImageNet. The accuracy and computational cost of the used Student and Teacher models for the three datasets are summarized in Table \ref{tbl:ranet_models}. For the Student, we employed the RANet's first classifier from the first scale with 0.316 ($\times 10^8$) FLOPs. For the Teacher, the last classifier from the last scale with 1.89 ($\times 10^8$) FLOPs was used. 
%the accuracy of 88.81\%
%the accuracy of 90.93\%
%Note that for faster training, we utilized the authors' code but used a batch size of 4096 and learning rate of 0.6, hence the results are slightly different than the ones reported in \cite{ranet}. 
Figure \ref{fig:ebjr_ranet} shows the corresponding adaptive inference results compared with the RANet baseline on CIFAR-10, CIFAR-100, and ImageNet. We observe that our method is orthogonal to RANet, and can improve it further.

\begin{table}
\centering
\fontsize{8}{10}\selectfont
\begin{tabular}[t]{p{1.5cm}cccccc}
\toprule
\multicolumn{1}{c}{ } & \multicolumn{2}{c}{\textbf{CIFAR-10}} & \multicolumn{2}{c}{\textbf{CIFAR-100}} & \multicolumn{2}{c}{\textbf{ImageNet}}\\ 
\cmidrule(l{3pt}r{3pt}){2-3} \cmidrule(l{3pt}r{3pt}){4-5} \cmidrule(l{3pt}r{3pt}){6-7} 
& \textbf{S} & \textbf{T} & \textbf{S} & \textbf{T} & \textbf{S} & \textbf{T} \\
\midrule
%\hspace{-5pt}\textbf{Depth} & 52 & 64 & 58 & 88 & 121 & 201\\
%\hspace{-5pt}\textbf{Growth Rate} & 6 & 12 & 6  & 8 & 12 & 32\\
\hspace{-5pt}\textbf{Accuracy} {\tiny{($\%$)}} & 91.18 & 93.61 & 67.28 & 74.73 & 56.18 & 71.69 \\
\hspace{-5pt}\textbf{FLOPs} {\tiny{($\times10^8$)}} & 0.3162 & 1.898 & 0.3166 & 1.9 & 3.36 & 33.62\\
\bottomrule
\end{tabular}
\vspace{-5pt}
\caption{\label{tbl:ranet_models}\small Details of the Student (\textbf{S}) and Teacher (\textbf{T}) EBJR (RANet) models for CIFAR-10, CIFAR-100, and ImageNet experiments.}
\end{table}

\vspace{-12pt}

\subsection{Alternative routing mechanisms: Softmax and Entropy}
In Section \ref{sec:experiments_ai_ic}, an ablation study with some experiments (Figure \ref{fig:ic_comparison_results_imagenet}-right) was presented to analyze the softmax and entropy scores as alternative means of analyzing the Student. Here, we study the mathematical connection of them with the energy score and their potential to solve the routing problem.

\subsubsection{Softmax-based Router}
\label{ssec:method_jr_confidence}
\vspace{-3pt}
%In other words, a good score that is able to tell us if the input data is considered in-distribution for our Student model or not (i.e., misclassified or out-of-distribution). 
%Mathematical connection between softmax and energy scores:

The softmax score for a classifier is expressed by:
\begin{equation}
\label{eq:softmax_score}
    \max_{y} p(y|\textbf{x}) = \max_{y} \frac{e^{S^c_y(\textbf{x})}}{\sum_i^C e^{S^c_i(\textbf{x})}} =  
    \frac{e^{S^c_{max}(\textbf{x})}}{\sum_i^C e^{S^c_i(\textbf{x})}}.
\end{equation}
% which can be reformulated as:
% \begin{equation}
% \label{eq:softmax_score2}
%     \begin{split}
%     \max_{y} p(y|\textbf{x}) = \frac{e^{S^c_{max}(\textbf{x})}}{\sum_i^C e^{S^c_i(\textbf{x})}}.
    % \\
    % = \frac{e^{S^c_{max}(\textbf{x})} / e^{S^c_{max}(\textbf{x})}}{\sum_i^C e^{S^c_i(\textbf{x})} / e^{S^c_{max}(\textbf{x})}}
    % \\
    % = \frac{1}{\sum_i^C e^{S^c_i(\textbf{x})-S^c_{max}(\textbf{x})}}.
%     \end{split}
% \end{equation}
%Log of softmax confidence score (based on Equation 4):
By taking the logarithm of both sides, 
we start to see the connection between the log of the softmax and the free energy score formulated in (\ref{eq:classifier_free_energy}):
%\vspace{-10pt}
\begin{equation}
    \begin{split}
    log \max_{y} p(y|\textbf{x}) = 
    log (e^{S^c_{max}(\textbf{x})}) - log \sum_i^C e^{S^c_i(\textbf{x})}
    = S^c_{max}(\textbf{x}) + F^c(\textbf{x};S^c),
    \end{split}
\end{equation}
where all logits are shifted by their maximum logit $S^c_{max}(x)$. Plugging in the energy term to (\ref{eq:classifier_log_density}) yields:
%For in-distribution samples, $S^{max}(x)$ tends to be higher, but $F(x;S)$ tends to be lower! In other words, $S^{max}(x)$ tends to be lower and $E(x;f)$ tends to be higher for mis-classified (OOD) samples. This will make the shifting a biased scoring function, where (unlike energy score) is not well-aligned with density $p(x)$. So, based on equation 7:
\begin{equation}
\label{eq:softmax_log_density}
     \begin{split}
        \hspace{-6pt} log \max_{y} p(y|\textbf{x}) = -log(p(\textbf{x})) + S^c_{max}(\textbf{x}) - log(Z^c).
     \end{split}
\end{equation}

%\vspace{-8pt}
It is observed that for the samples with high likelihood of being in the Student's distribution, the free energy goes lower, but the maximum logit tends to go higher. Due to this shifting, unlike the energy score, the softmax confidence score is not well-aligned with the probability density $p(\textbf{x})$. %In other words, the normalization (the right two terms) in the above equation can affect OOD detection (unlike energy score). 
As a result, the confidence score is less reliable for our Router to analyze the performance of the Student.

\subsubsection{Entropy-based Router}
The entropy score is a measure of the randomness in the information being processed, and is calculated as follows:
\begin{equation}
    \begin{split}
    %H(\textbf{x};S^c)=-\sum_i^C p(y|\textbf{x}).log (p(y|\textbf{x})) \\
    H(\textbf{x};S^c)=-\sum_i^C S^c_i.log (S^c_i),
    %= -\sum_i^C \left ( \frac{e^{S^c_i(\textbf{x})}}{\sum_i^C e^{S^c_i(\textbf{x})}} \right).log \left( \frac{e^{S^c_c(\textbf{x})}}{\sum_i^C e^{S^c_i(\textbf{x})}} \right)
    \\
    %= -\sum_i^C \left ( \frac{e^{S^c_i(\textbf{x})}}{Z} \right).log \left( \frac{e^{S^c_i(\textbf{x})}}{Z} \right)
    %H(\textbf{x};S) = -\sum_i^C f_i(\textbf{x}).log (f_i(\textbf{x}))
    \end{split}
\end{equation}
where $S^c_i(\textbf{x})$ is the probability (logit) corresponding to the $i$-th class label. 

Let $U$ be the internal energy (i.e., the expectation value of the energy function \cite{oh2020entropy}), defined by:
\begin{equation}
    \begin{split}
    U(\textbf{x};S^c) = \sum_i^C E(\textbf{x},i) S^c_i.
    %U(\textbf{x};S^c) = \sum_i^C E(\textbf{x},i) p(i|\textbf{x}) \\
    %= - \sum_i^C S^c_i(\textbf{x}) p(i|\textbf{x}) \\
    %= - \sum_i^C S^c_i(\textbf{x}) \left( \frac{e^{S^c_i(\textbf{x})}}{\sum_j^C e^{S^c_j(\textbf{x})}} \right) \\
    %= - \sum_i^C \left( \frac{S^c_i(\textbf{x}) . e^{S^c_i(\textbf{x})}}{Z} \right) \\
    %= - \frac{\sum_i^C \left( S^c_i(\textbf{x}) . e^{S^c_i(\textbf{x})} \right)}{Z}
    \end{split}
\end{equation}

According to \cite{oh2020entropy}, the entropy can be defined in terms of the internal and free energy functions as:
\begin{equation}
    H(\textbf{x};S^c) = U(\textbf{x};S^c) - F(\textbf{x};S^c),
\end{equation}
where all logits are shifted by the internal energy $U$. 
 
Substituting the free energy term from (\ref{eq:classifier_log_density}) yields:
\begin{equation}
 \begin{split}
    H(\textbf{x};S^c) = log(p(\textbf{x})) + U(\textbf{x};S^c) + log(Z^c),
 \end{split}
\end{equation}
which shows that, due to the shifting caused by the internal energy, the entropy score is not reliably aligned with the probability density $p(\textbf{x})$. Thus, it is a less suitable mechanism to be used as a routing mechanism in our Router, as opposed to the energy score.

\subsection{Imbalance in class distributions}
\begin{figure*}
    \centering
    \includegraphics[width=0.49\linewidth]{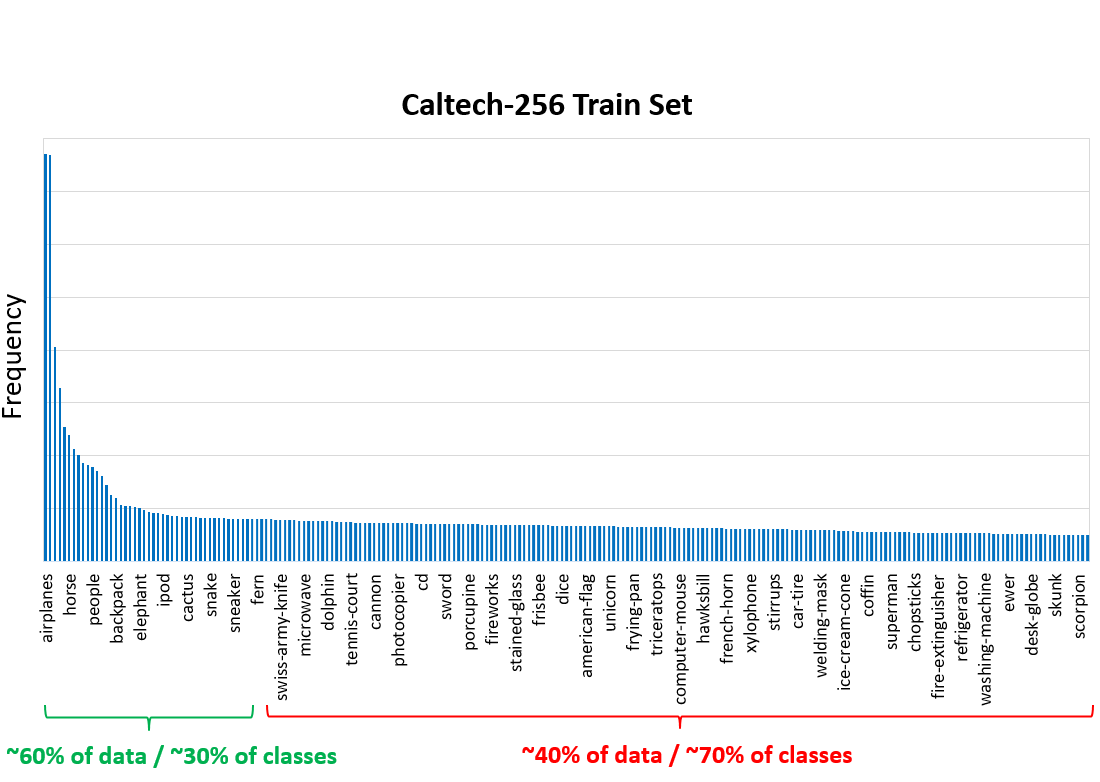}
    \includegraphics[width=0.49\linewidth]{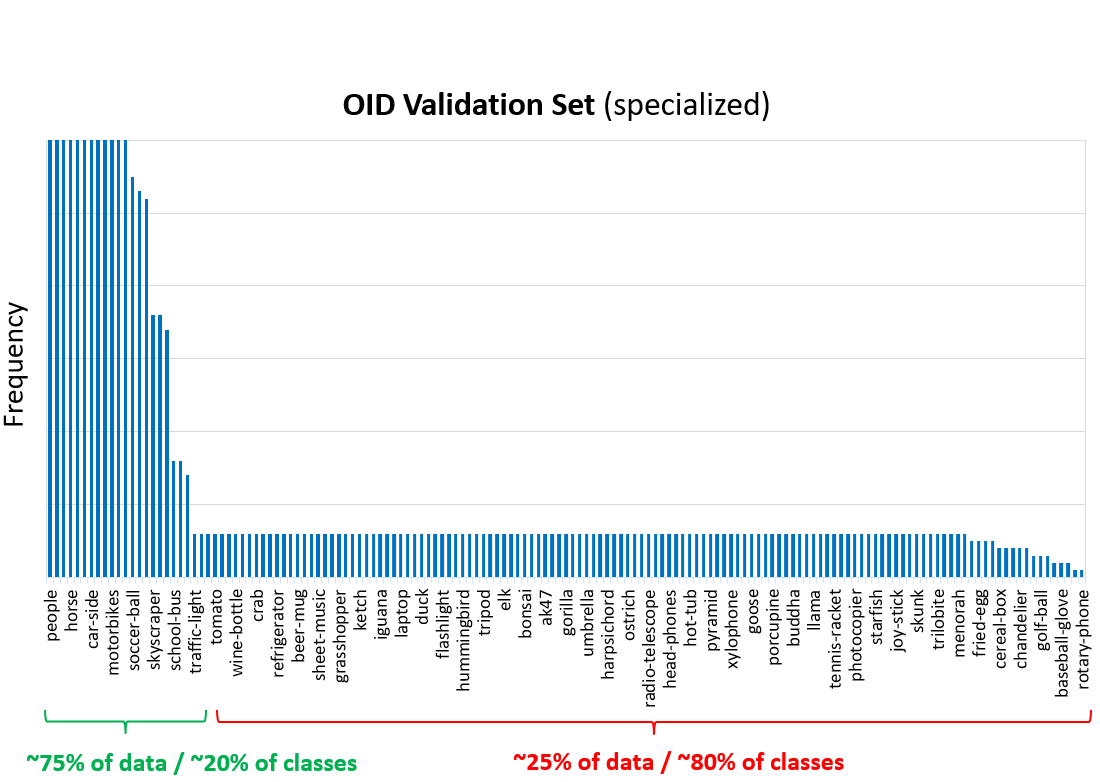}
    \vspace{-4pt}
    \caption{Data and class distributions over the Caltech-256 train set and OID validation set (for image classification with 256 labels). Due to the limited space, only a subset of class names are shown on the X axis for better visualization.}
    \label{fig:caltech_oid_dist}
\end{figure*}

\begin{figure*}
    \centering
    \includegraphics[width=0.49\linewidth]{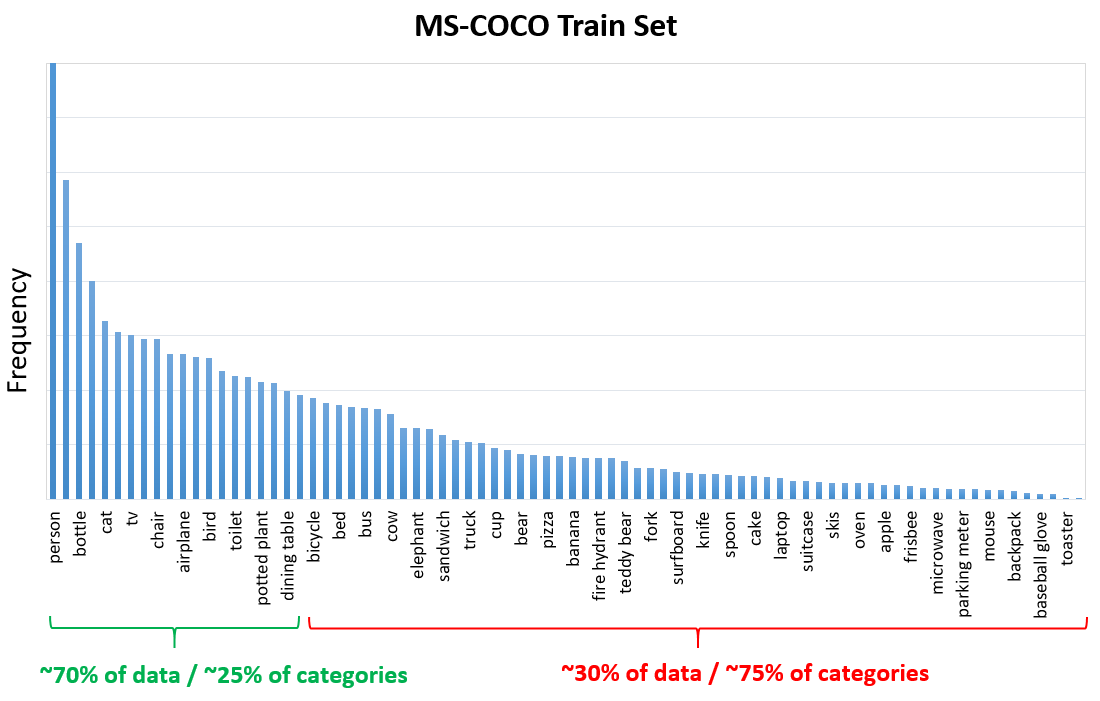}
    \includegraphics[width=0.49\linewidth]{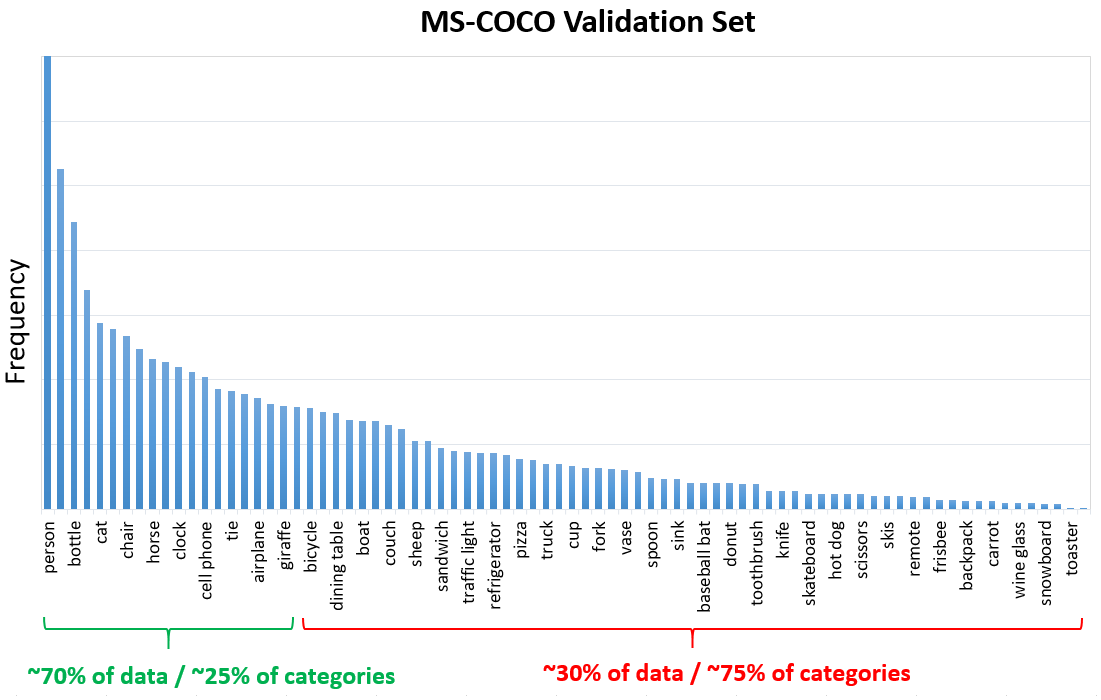}
    \vspace{-4pt}
    \caption{Data and category distributions over the OID train set and MS-COCO validation set (for object detection with 80 categories). Due to the limited space, only a subset of category names are shown on the X axis for better visualization.}
    \label{fig:coco_oid_dist}
\end{figure*}

In Section \ref{ssec:method_jr_specialized}, it was mentioned that in many practical applications, training or testing datasets are imbalanced. For example, consider a cloud inference API, which receives images as input, and most of the input images belong to a limited number of popular classes or categories. This motivated the specialized EBJR case. We studied the class distribution for the Caltech-256, OID, and MS-COCO datasets in Figures \ref{fig:caltech_oid_dist} and \ref{fig:coco_oid_dist}, and the statistics confirm our intuition. 

\begin{figure*}
    \centering
    \includegraphics[width=0.49\linewidth]{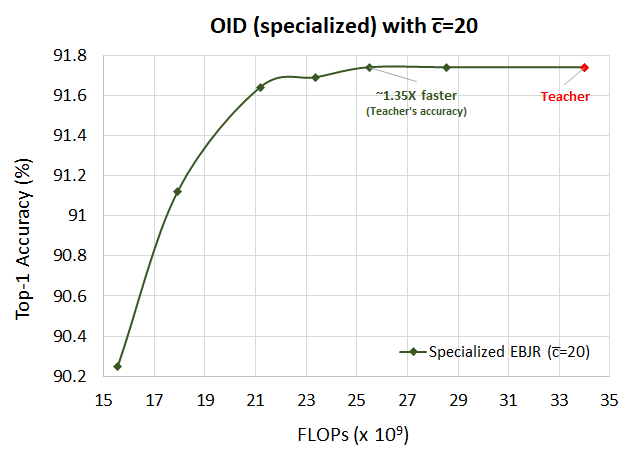}
    \includegraphics[width=0.49\linewidth]{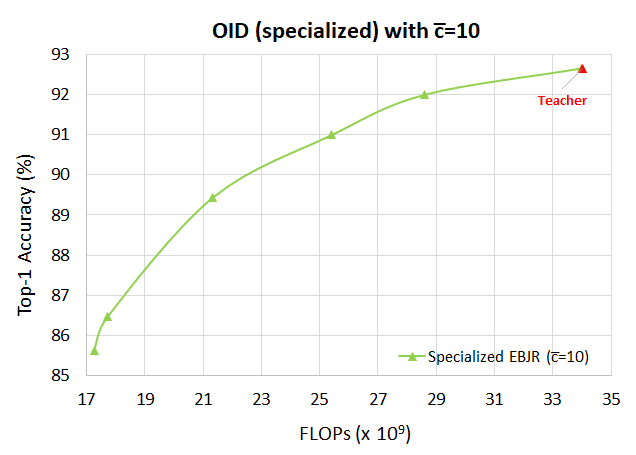}
    \vspace{-10pt}
    \caption{Adaptive inference results of the specialized EBJR with $\Bar{C}=20$ and $\Bar{C}=10$ on OID dataset.}
    \label{fig:caltech256-different-topn}
\end{figure*}

\subsection{More results on the specialized EBJR} 

Figure \ref{fig:caltech256-different-topn} shows the adaptive inference results for the specialized EBJR case. This figure shows the top-1 classification accuracy of joint models when top-$\bar{C}$=10 or 20 popular classes are used. For top-$\bar{C}$=10, we choose the top-10 class labels with the most number of samples in the OID training set, and for testing, we randomly select a new set of size 1.7K from the OID validation set, where 70\% of the data have the top-10 of the class labels. For top-$\bar{C}$=20, the size of the corresponding randomly selected validation set is 2K, where 75\% of the samples belong to the top-20 labels. It should be noted that the Teacher accuracies over the top-10 and top-20 validation sets are not the same because the validation sets are not identical (different data/label distribution).

It is observed from Figure \ref{fig:caltech256-different-topn} that top-$\bar{C}$=20 results in a better overall performance, and can achieve the same accuracy as the teacher but with 1.35$\times$ faster inference. Reducing the number of classes to 10 will make the performance worse, where almost no speed-up can be achieved compared to the Teacher. This suggests that limiting the majority of popular categories to a very low number of classes may hurt the performance.

\subsection{More insights on inequality (\ref{eq:energy_gap})}
%For the specialized EBJR \mo{this is not for the specialized EBJR, it is the general case for the free energy}, 
The free energy of the Student $S^c(\textbf{x}): \mathbb{R}^D \rightarrow \mathbb{R}^C$ in (\ref{eq:classifier_free_energy}) can be broken into the logarithm of two terms as:
\begin{equation}
\begin{split}
    F^c(\textbf{x};S^c) = %-log \sum_i^C e^{S^c_i(\textbf{x})} \\
    - log \left( e^{S^c_y} + \sum_i^{\Bar{C}} e^{S^c_i(\textbf{x})} \right),
\end{split}
\end{equation}
where $C = \Bar{C} + 1$ and $y \in \{\Bar{C} + 1\}$. Factoring out the term $e^{S^c_y}$ from inside the logarithm yields:
\begin{equation}
\begin{split}
    F^c(\textbf{x};S^c) = - log(e^{S^c_y}) - log \left(1 + \frac{\sum_i^{\Bar{C}} e^{S^c_i(\textbf{x})}}{e^{S^c_y}} \right).
\end{split}
\end{equation}

By denoting the second term as $\hat{F^c}(\textbf{x};S^c)$, we will have:
\begin{equation}
\label{eq:free_energy_new}
    F^c(\textbf{x};S^c) = - \big(S^c_y + \hat{F^c}(\textbf{x};S^c)\big).
\end{equation}

\begin{table*}
\centering
\fontsize{8}{10}\selectfont
\begin{tabular}[t]{p{1.5cm}cccccc}
\toprule
\multicolumn{1}{c}{ } & 
\multicolumn{3}{c}{\textbf{Caltech-256}} & 
\multicolumn{3}{c}{\textbf{OID}}\\ 
\cmidrule(l{3pt}r{3pt}){2-4} \cmidrule(l{3pt}r{3pt}){5-7}
& \textbf{S} & \textbf{S$^*$} & \textbf{T} & \textbf{S} & \textbf{S$^*$} & \textbf{T} \\
\midrule
%\hspace{-5pt}\textbf{Supervised} & 83.16 & & & \\
%\hspace{-5pt}\textbf{Unsupervised} & 70.71 & N/A & 58 & N/A\\
\hspace{-5pt}\textbf{Accuracy} {\tiny{($\%$)}} & 83.16 & 70.71 & 89.87 & 75.0 & 74.1 & 86.64 \\
\hspace{-5pt}\textbf{FLOPs} {\tiny{($\times10^9$)}} & 5.4 & 5.4 & 34.0 & 5.4 & 5.4 & 34.0\\
\bottomrule
\end{tabular}
%\vspace{-10pt}
\caption{\label{tbl:caltech_models} \small The performance of the supervised and unsupervised Student (respectively denoted by \textbf{S} and \textbf{S$^*$}) and the Teacher (\textbf{T}) on Caltech-256 and OID validation sets.}
\end{table*}

Let $(x,y)$ be in-distribution and $(\Bar{x},\Bar{y})$ be out-of-distribution samples, where $y \in [1,\Bar{C}]$ and $\Bar{y} \notin [1,\Bar{C}]$. Based on (\ref{eq:free_energy_new}), the inequality (\ref{eq:energy_gap}) can be reformulated as:
\begin{equation}
\scriptsize
\begin{split}
%\underbrace{(x + 2)^3}_\text{text 1}
    | \underbrace{\Bar{F}^c(\textbf{x};\Bar{S}^c)}_\text{decrease} - \underbrace{\Bar{F}^c(\Bar{\textbf{x}};\Bar{S}^c)}_\text{increase} | > | \big( \underbrace{S^c_{\Bar{y}}}_\text{decrease} +\underbrace{\hat{F^c}(\Bar{\textbf{x}};S^c) }_\text{increase} \big) - \big( \underbrace{S^c_y}_\text{increase} +\underbrace{\hat{F^c}(\textbf{x};S^c)}_\text{decrease} \big) |,
\end{split}
\label{sup:eq:gaps}
\end{equation}
where the followings can be observed for the left side of this inequality:
\begin{itemize}
    \item 
    Since $(x,y)$ is an in-distribution sample (high likelihood) and also $y \in [1,\Bar{C}]$, the free energy $\Bar{F}^c(\textbf{x};\Bar{S}^c)$ tends to be lower.
    \item
    Since $(\Bar{x},\Bar{y})$ is an out-of-distribution sample (low likelihood) and also $\Bar{y} \notin [1,\Bar{C}]$, the free energy $\Bar{F}^c(\Bar{\textbf{x}};\Bar{S}^c)$ tends to be higher. 
\end{itemize}
And for the right side:
\begin{itemize}
    \item
    For $(x,y)$, the $y$-th logit value $S^c_y$ tends to increase (high likelihood), which makes $\hat{F^c}(\textbf{x};S^c)$ decrease.
    \item
    For $(\Bar{x},\Bar{y})$, the $\Bar{y}$-th logit value $S^c_{\Bar{y}}$ tends to decrease (low likelihood), which makes $\hat{F^c}(\Bar{\textbf{x}};S^c)$ to increase.
\end{itemize}

The two terms on the left side tend to go in the opposite directions, thereby enlarging the energy difference. On the other hand, the two terms on the right side of (\ref{sup:eq:gaps}) do not show a similar behaviour, and thus their gap does not necessarily increase.

\subsection{Unsupervised EBJR} 
\label{ssec:method_jr_unsupervised}
So far, it was assumed that the Teacher and Student already-trained models are given, and with those we created a joint inference bundle model. This assumption may not always be true.
Suppose we have a large model that is highly accurate, but also very slow. In addition, no dataset with ground truth labels is available to train a small and fast Student model. In this scenario, in order to achieve an efficient joint reasoning model, we can effectively distill the Teacher model to a small and fast Student architecture, in a completely unsupervised manner. Unsupervised knowledge distillation is an emerging technique for leveraging the abundance of unlabeled data for label-free model training. Our framework is flexible in that it can organically incorporate the unsupervised distillation.

The most straight-forward application of unsupervised EBJR is for cloud services, which include very large models for different machine learning tasks served through cloud APIs. Such inference services can be replaced by our EBJR architecture in which a side Student model is created for each large model. In this case, there is no need for re-training the large models nor acquiring data labels. By replacing the current large models behind the APIs with the joint reasoning equivalent, a speed-up gain can be achieved without a considerable loss in accuracy. For the classification problem, as an example, the commonly used cross-entropy loss function for training the Student is given by:
%\vspace{-6pt}
\begin{equation}
    CE = - \sum_i^C \tau_i~log(S^c_i(\textbf{x})) ~~~\text{with}~~~ \tau_i = T^c(\textbf{x}),
\end{equation}

%\vspace{-8pt}
\noindent where the pseudo-labels generated by the Teacher model are utilized as the targets (denoted by $\tau_i$) in the loss function. 
%\mo{do we need to add something for object detection as well?} -> will try to add it.

\begin{figure*}
    \centering
    \begin{subfigure}{0.51\textwidth}
        \centering
        \includegraphics[width=0.99\linewidth]{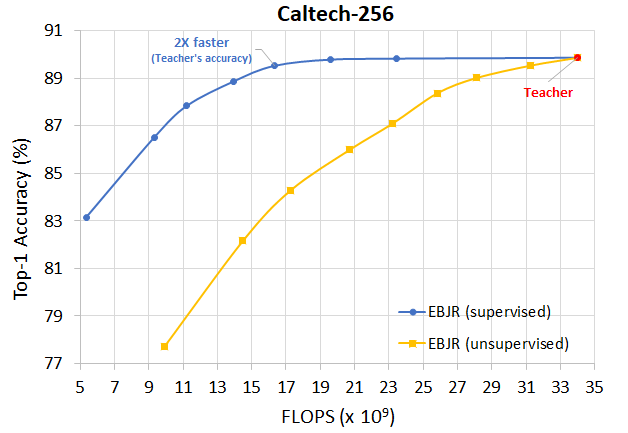}
        % \caption{}
        \label{fig:caltech256-oid-comparison_caltech256}
    \end{subfigure}
    \hspace{-13pt}
    \begin{subfigure}{0.51\textwidth}
        \centering
        \includegraphics[width=0.99\linewidth]{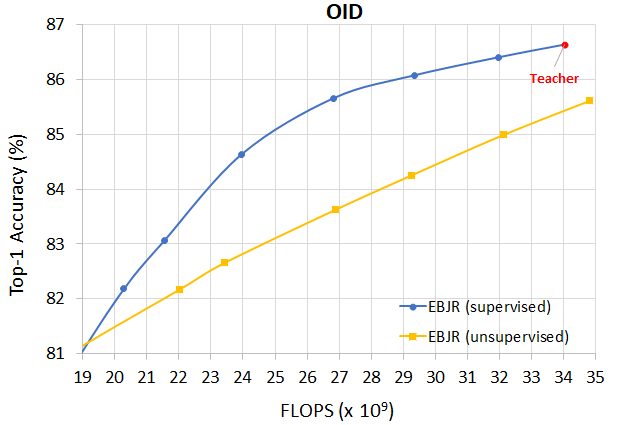}
        % \caption{}
        \label{fig:caltech256-oid-comparison_oid}
    \end{subfigure}
    \vspace{-15pt}
    \caption{\small The performance of supervised and unsupervised EBJR on Caltech-256 and OID validation sets.}
    \label{fig:caltech256-oid-comparison}
\end{figure*}

\subsubsection{Experimental results - image classification} 
\label{ssec:experiments_jr_unsupervised}

In this section, we study the performance of the unsupervised version of our method (see Section \ref{ssec:method_jr_unsupervised}). To this end, we perform unsupervised distillation on the Student, using a set of unlabeled examples, which are passed to the Teacher to obtain pseudo-labels. The Student is then trained purely with these pseudo-labels. In this experiment, we use a ResNet-152 pre-trained on the Caltech-256 training set (22K examples in 256 classes) as the Teacher. The Student is a ResNet-18 trained with a 56K unlabeled random subset of OID. 

For testing, we evaluated our approach on two validation sets including Caltech-256 (7.8K images) and a subset of OID validation set (12K images). The accuracies and computational costs of the Student (supervised and unsupervised) and Teacher on both validations sets are reported in Table \ref{tbl:caltech_models}. Note that we study these two validations sets since they can both be valid measures depending on the target application. One represents the case when a user provides a large Teacher model with some validation data for which the joint model needs to attain a high accuracy. The other represents the case where a user provides a large Teacher model, and the joint model is supposed to work well for data that hit the cloud API, which are similar to the unlabeled data used to train the unsupervised joint model.

Figure \ref{fig:caltech256-oid-comparison} presents the adaptive inference results with the unsupervised EBJR and also its comparison with the supervised case on both validation sets. For the supervised EBJR, the Student is trained on Caltech-256 (similar to the Teacher). As observed in Figure \ref{fig:caltech256-oid-comparison}-left, the unsupervised EBJR does not perform as well as the supervised case, which is because the distributions of the training and testing sets are different (OID vs. Caltech-256). However, when evaluated on the OID validation set, which follows the same distribution with which the Student is trained, a better performance is achieved (Figure \ref{fig:caltech256-oid-comparison}-right). 

%\vspace{-2pt}
It is shown in \cite{xie2020self, sohn2020simple, zoph2020rethinking} that using large amounts of unlabeled data for pseudo-label self-training can achieve results even higher than the supervised models. In agreement with this observation, we will see later in this section that the performance of the unsupervised joint model tends to improve if larger amounts of unlabeled data are used. In some cases, it may even surpass the performance of the supervised model (see Figure \ref{fig:od_comparison_unsup}). That being said, the results in Figure \ref{fig:caltech256-oid-comparison} are excellent for the supervised case, and still very promising for the unsupervised case, as the later is not using any labels for training the joint model.

\subsubsection{Experimental results - object detection} 
\label{ssec:experiments_jr_unsupervised_od}

We also analyze the performance of the unsupervised variant of EBJR on the task of object detection on the MS-COCO dataset, where we employ the EfficientDet-D0 and EfficientDet-D4 architectures \cite{efficientdet} for the Student and Teacher, respectively. For the unsupervised EBJR, the OID training set is used as the unlabeled set.

\begin{figure*}
    \centering
    \includegraphics[width=0.51\linewidth]{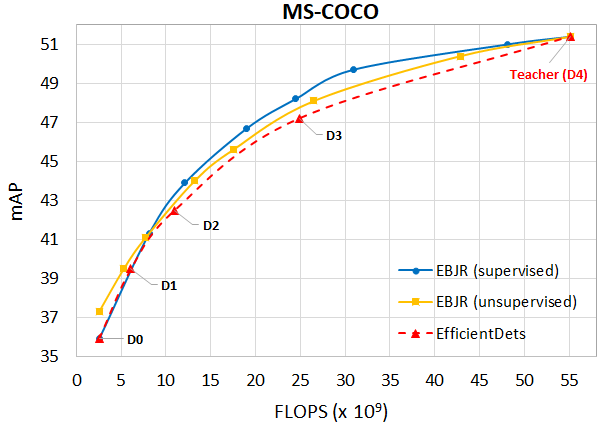}
    \vspace{-6pt}
    \caption{The adaptive inference performance of the supervised and unsupervised EBJR for object detection on MS-COCO (compared with the EfficientDet models \cite{efficientdet}).}
    \label{fig:od_comparison_unsup}
\end{figure*}

\begin{table*}
\centering
\fontsize{8}{10}\selectfont
\begin{tabular}[t]{p{1.6cm}p{1.25cm}p{0.64cm}p{0.5cm}p{0.5cm}p{1.25cm}}
\toprule
\multicolumn{1}{c}{ } & \multicolumn{4}{c}{\textbf{Student}} & \multicolumn{1}{c}{\textbf{Teacher}}\\ 
\cmidrule(l{3pt}r{3pt}){2-5} \cmidrule(l{3pt}r{3pt}){6-6} 
\hspace{-7pt} \textbf{Mode} & Supervised & \multicolumn{3}{c}{\hspace{-8pt} Unsupervised} & Supervised\\
\cmidrule(l{3pt}r{3pt}){2-2} \cmidrule(l{3pt}r{3pt}){3-5} \cmidrule(l{3pt}r{3pt}){6-6}
\hspace{-7pt} \textbf{Train-set size} & 118K & 160K & \hspace{-6pt} 320K & \hspace{-6pt} \textbf{1.7M} & 118K\\
\hspace{-7pt} \textbf{mAP} & 0.359 & 0.329 & \hspace{-6pt} 0.350 & \hspace{-6pt} \textbf{0.373} & 0.514\\
\cmidrule(l{3pt}r{3pt}){2-5}
\hspace{-7pt} \textbf{FLOPs} {\tiny{($\times 10^9$)}} & \multicolumn{4}{c}{2.54} & 55.2\\
\bottomrule
\end{tabular}
\vspace{-4pt}
\caption{\label{tbl:od_models}\small The performance of the Teacher and the supervised and unsupervised Student with different train-set sizes on MS-COCO.}
\end{table*}

Table \ref{tbl:od_models} reports the performance of the Student model trained in the supervised and unsupervised settings, compared to the Teacher. For the unsupervised case, we tested different amounts of unlabeled data from OID. We observe that when sufficient unlabeled data (e.g., 1.7M in Table \ref{tbl:od_models}) are provided, the unsupervised Student can perform even better than the supervised one.

Moreover, Figure \ref{fig:od_comparison_unsup} shows the adaptive inference results for both supervised and unsupervised (with 1.7M samples from OID) cases compared to the EfficientDet models (D0, D1, D2, D3, and D4). Both supervised and unsupervised EBJR outperform the standard EfficientDet models.